\pdfoutput=1

\documentclass[11pt]{article}

\PassOptionsToPackage{dvipsnames,usename}{xcolor}
\usepackage{acl}

\usepackage{times}
\usepackage{latexsym}

\usepackage[T1]{fontenc}

\usepackage[utf8]{inputenc}

\usepackage{microtype}

\usepackage{inconsolata}
\usepackage{graphicx}
\usepackage{amsmath}
\usepackage{amssymb}
\usepackage{amsthm}
\usepackage{amsfonts}
\usepackage{bbm}
\usepackage{xspace}
\usepackage{pbox}
\usepackage{booktabs}
\usepackage{hyperref}
\hypersetup{
    colorlinks=true,
    linkcolor=darkblue,
    filecolor=darkblue,      
    urlcolor=darkblue,
}

\usepackage{makecell}
\usepackage{scalerel}
\usepackage{array}

\usepackage{multirow}
\usepackage{caption}
\usepackage{subcaption}
\usepackage{xfrac}

\usepackage{cleveref}
\crefname{section}{\S}{\S\S}
\Crefname{section}{\S}{\S\S}
\crefformat{section}{\S#2#1#3}
\crefname{figure}{Fig.}{Fig.}
\crefname{alg}{Alg.}{Alg.}
\crefname{line}{line}{lines}
\crefname{appendix}{App.}{}
\crefname{equation}{Eq.}{Eqs.}
\crefname{table}{Table}{Tables}
\crefname{proposition}{Proposition}{Propositions}

\newtheorem{takeaway}{Takeaway}

\usepackage[textsize=tiny]{todonotes}
\newcommand{\note}[4][]{}

\newcommand{\tiago}[2][]{\note[#1]{tiago}{cyan!40}{#2}}

\newcommand{\clara}[2][]{\note[#1]{Clara}{orange!40}{#2}}

\newcommand{\response}[1]{\vspace{3pt}\hrule\vspace{3pt}\textbf{#1:}}
\newcommand{\anton}[2][]{\note[#1]{anton}{green}{#2}}
\newcommand{\Anton}[2][]{\anton[inline,#1]{#2}}

\newcommand{\english}{\mathrm{EN}}

\newcommand{\french}{\mathrm{FR}}

\newcommand{\upSymbol}{$\boldmath \Rsh$}
\newcommand{\downSymbol}{\rotatebox[origin=c]{180}{$\boldmath \Lsh$}}
\newcommand{\staySymbol}{$\boldmath \rightarrow$}
\newcommand{\stepUp}[2]{#1 $\boldmath \Rsh$ #2}
\newcommand{\stepDown}[2]{#1 \rotatebox[origin=c]{180}{$\boldmath \Lsh$} #2}

\newcommand{\defn}[1]{\textbf{#1}}

\newcommand{\saveForCr}[1]{}

\title{
The Role of Language Imbalance in Cross-lingual Generalisation:\\Insights from Cloned Language Experiments
}

\newcommand*\samethanks[1][\value{footnote}]{\color{darkblue} \footnotemark[#1]}

\newcommand{\emailadress}[1]{\texttt{#1}}

\newcommand{\ethz}{1}
\newcommand{\ethai}{3}
\newcommand{\barilan}{2}

\author{Anton Sch\"afer$^\ethz$, Shauli Ravfogel$^{\barilan}$, Thomas Hofmann$^\ethz$, Tiago Pimentel$^{\ethz,}$\thanks{Joint supervision.}, Imanol Schlag,$^{\ethai,}$\samethanks\\
  $^\ethz$ETH Z\"urich, $^{\barilan}$Bar-Ilan University, $^\ethai$ETH AI Center \\
   \emailadress{scanton@ethz.ch}, \emailadress{shauli.ravfogel@gmail.com}, \\
   \{\emailadress{thomas.hofmann}, \emailadress{imanol.schlag}, \emailadress{tiago.pimentel}\}\texttt{@inf.ethz.ch}
}

\begin{document}
\maketitle
\begin{abstract}
Multilinguality is crucial for extending recent advancements in language modelling to diverse linguistic communities. 
To maintain high performance while representing multiple languages, multilingual models ideally align representations, allowing what is learned in one language to generalise to others.
Prior research has emphasised the importance of parallel data and shared vocabulary elements as key factors for such alignment. 
In this study, we investigate an unintuitive novel driver of cross-lingual generalisation: language \emph{imbalance}. 
In controlled experiments on perfectly equivalent cloned languages, we observe that the existence of a predominant language during training boosts the performance of less frequent languages and leads to stronger alignment of model representations\clara{perhaps be more explicit here about which type of representations? like 'token' representations or something} across languages.
Furthermore, we find that this trend is amplified with scale: with large enough models or long enough training, we observe that bilingual training data with a \sfrac{90}{10} language split yields better performance on both languages than a balanced \sfrac{50}{50} split.
Building on these insights, we design training schemes that can improve performance in all cloned languages, even without altering the training data. 
As we extend our analysis to real languages, we find that infrequent languages still benefit from frequent ones, yet whether language imbalance causes cross-lingual generalisation there is not conclusive.\looseness=-1

 \vspace{.3em}
 \hspace{.5em}\includegraphics[width=1.25em,height=1.25em]{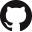}{\hspace{.75em}\parbox{\dimexpr\linewidth-2\fboxsep-2\fboxrule}{\vspace{-5pt} \href{https://github.com/antonschafer/xling-imbalance/}{antonschafer/xling-imbalance}}}
\end{abstract}

\newcommand{\sectionvspace}{\vspace{-4pt}}
\newcommand{\subsectionvspace}{\vspace{-2pt}}
\setlength{\abovecaptionskip}{3pt}

\section{Introduction}
\sectionvspace

\newcommand{\mask}{\texttt{mask}\xspace}
\newcommand{\bos}{\texttt{bos}\xspace}
\newcommand{\wordform}[1]{\textit{#1}\xspace}

In recent years, autoregressive language models (LMs) pretrained on massive text corpora have advanced the state of the art in NLP tasks across the board \citep{brown2020gpt3, touvron2023llama,touvron2023llama2, kopf2023openassistant}.
While most of the leading models are trained on English texts, multilingual capabilities are crucial to make these advances accessible to a broader user base with diverse linguistic backgrounds.
Ideally, data in one language should improve these multilingual models' performance in others.
Such multilingual models should thus display \defn{cross-lingual generalisation}: by reusing circuits  \citep{cammarata2020thread,elhage2021mathematical} and aligning their internal representations across languages, they may generalise concepts learned in a language to another.\looseness=-1
\footnote{A circuit is typically defined as a subgraph of a neural network which performs a specific computation. E.g., a circuit could be responsible for computing ``greater than'' comparisons between numbers in English sentences \citep{hanna2024does}. If representations are aligned across languages (in terms of how they encode, e.g., numbers) and circuits are reused, a model should be able to apply what it learns in one language (e.g., ``greater than'' comparisons in English) to perform similar computations in another language (e.g., French).
\looseness=-1
}
\looseness=-1

How can such cross-lingual generalisation be achieved?
This has been a focus of much prior work.
One previously identified driver
of cross-lingual generalisation is \defn{parallel training data};
empirical evidence shows that training the model on either parallel sentence pairs \citep{lample2019cross} or on corpora which are comparable across languages \citep{dufter2020identifying} improves generalisation.
Another driver of cross-lingual generalisation is the availability of \defn{anchor points}, i.e., vocabulary elements that are shared between languages;
these can be naturally occurring subwords \citep[e.g., \wordform{computer} in English and \wordform{computador} in Portuguese may share the subword \wordform{comp};][]{pires2019multilingual, wu2019beto}, shared special tokens \citep[e.g., \mask or \bos symbols;][]{dufter2020identifying}, or even artificially inserted ``code-switching'' augmentations \citep{conneau2020emerging, reid2022paradise, wang2019cross, feng2022toward}.
Beyond these two drivers, \defn{limited model capacity} has been found to improve generalisation
by \citet{dufter2020identifying}, but to constrain multilingual capabilities by \citet{chang2024when}.\looseness=-1

In this work, we identify a surprising new factor that can boost cross-lingual generalisation abilities: \defn{language imbalance}.
We first conduct experiments in a synthetic setting with perfectly equivalent cloned languages; this allows us to investigate LMs' generalisation abilities in isolation from the effects of languages' dissimilarities, giving us a rough upper bound on the generalisation we should expect to see between real language pairs.
In this cloned language setting, we find that having a dominant 
main language improves generalisation, significantly boosting the performance of less frequent languages.
Furthermore, we find that this effect becomes stronger when we either increase our model's size or when we train it for longer.
Based on these insights, we design training curricula that improve performance in all cloned languages without any modifications to the training data.
In the second part of our paper, we investigate to what extent our insights transfer to real language pairs. 
While we find that lower resource
languages typically do benefit from higher resource ones, the impact of language imbalance on 
cross-lingual generalisation is much less clear in this more realistic setting. 
Overall, our results suggest an interesting attribute of model training dynamics: in some settings, having a main language can lead model components to be shared across languages.
\looseness=-1

\vspace{-5pt}
\section{Cross-lingual Generalisation}
\sectionvspace
\vspace{-2pt}

\newcommand{\origlabel}{\scalebox{.5}{\text{orig}}}
\newcommand{\multilabel}{\scalebox{.5}{\text{multi}}}
\newcommand{\lang}{L}
\newcommand{\langorig}{L_{\origlabel}}
\newcommand{\dataorig}{\dataset_{\origlabel}}
\newcommand{\datamulti}{\dataset_{\multilabel}}
\newcommand{\vocab}{\Sigma}
\newcommand{\word}{w}
\newcommand{\words}{\boldsymbol{w}}
\newcommand{\wordsorig}{\boldsymbol{w}_{\origlabel}}

\newcommand{\cloneequiv}{\stackrel{\scriptscriptstyle\circ}{=}}
\newcommand{\btheta}{\boldsymbol{\theta}}
\newcommand{\model}{p_{\btheta}}

\newcommand{\langa}{\mathrm{A}}
\newcommand{\langb}{\mathrm{B}}
\newcommand{\multi}{\mathrm{multi}}
\newcommand{\dataset}{\mathcal{D}}

While natural languages differ widely in their typological properties, any pair of languages will share at least a few grammatical and syntactic patterns.
Further, as their semantics reflect the underlying processes of our world, language pairs should also have similarities in the types of messages their users typically convey. 
Intuitively, this suggests that 
what is learned about a language $\lang_\langa$ 
should be useful to model another language $\lang_\langb$, and vice versa.
The extent of such generalisation depends not only on how similar
the two languages are, but also on the employed learning algorithm.
We analyse such generalisation here, with a focus on how language imbalance influences multilingual LMs.\looseness=-1

Intuitively, if a model generalises well across languages, it should achieve better performance in each language (in terms of, e.g., perplexity) than a monolingual model trained on the same data.
Concretely, a model trained on a multilingual dataset $\dataset_\multi = \dataset_\langa \cup \dataset_\langb$ containing languages $\lang_\langa$ and $\lang_\langb$ should perform better than monolingual models trained only on $\dataset_\langa$ or $\dataset_\langb$.
This becomes clear when using definitions from information theory: $\dataset_\multi$ contains at least as much information about $\lang_\langa$ as $\dataset_\langa$.
However, such a multilingual model could also perform worse.
This could happen, for instance, if the data from different languages interfere
with each other during optimisation through conflicting gradient update directions \citep{wang2020negative}. 
It could also happen if the model has limited capacity: the multilingual model has to represent many languages, which intuitively requires more capacity than a single one, even if some parameters are shared across them \citep{conneau-etal-2020-unsupervised,pfeiffer-etal-2022-lifting}.\looseness=-1

In an attempt to make models better across many languages, many multilingual models these days are trained on somewhat balanced data \citep{workshop2023bloom, faysse2024croissantllm}.\tiago{If there's time, add at least one more example here?}
In some of these cases, low-resource languages are upsampled to improve their performance under the model.
As mentioned above, however, while balancing languages' appearance in a model's training set should intuitively improve performance, this is not necessarily true.
In fact, (and perhaps surprisingly) some recent large language models trained in mostly English-focused settings perform reasonably well in a large sample of languages \citep{ahia-etal-2023-languages, blevins2022language, briakou2023searching}.
These models' training data is typically highly imbalanced, with only a small fraction being composed of ``non-English'' languages.
It is thus unclear whether multilingual models indeed benefit from training on datasets with balanced languages \citep{ye2023language}.\looseness=-1

In smaller training scales, the benefits of multilingual training are better understood.
In general, it has been found that low-resource languages tend to benefit from data in higher-resource languages, but high-resource languages benefit much less from each other \citep{conneau-etal-2020-unsupervised, chang2024when}.
It is, however, unclear what causes cross-lingual generalisation in this case.
Is the model in fact able to generalise better in the imbalanced setting?
Or does the model generalise equally well in the balanced case, but its capacity bottlenecks performance in higher-resource languages, stopping us from observing performance gains?\looseness=-1

We investigate the role of language imbalance in cross-lingual generalisation here.\anton{cite \cite{wei2023polylm}}
Notably, \citet{wendler2024llamas} recently showed that LMs seem to perform internal computations in an abstract ``concept space'' which is closest to their main language (English in this case); representations are then mapped back into the input language only in the models' final layers.
\citet{alabi2024hidden} observe a similar trend when using language adapters.\looseness=-1

\section{Experimental Setup}
\sectionvspace

In this section, we provide a brief overview over models, data, and metrics used; for more details, see \cref{app:setup}. Our code will be made available on GitHub.
All of our experiments use GPT-2-style decoder-only transformers \cite{radford2019language}. We base our implementation on the Languini Kitchen codebase \cite{stanic2023languini}, and unless otherwise noted, we use the \texttt{gpt-small} configuration with 85M non-embedding parameters, training on 1.2B tokens of English or French books.
 We use separate tokenisers for English and French. For some of our experiments, we treat their vocabularies as \defn{disjoint} and do not merge them. If we merge subwords that occur in both vocabularies, we make this clear with the label \defn{anchored}.

\newcommand{\ppl}{\ensuremath{\mathrm{PPL}}\xspace} 
\newcommand{\monotok}{\ensuremath{\mathrm{MLTE}}\xspace} 
\newcommand{\mlpe}{\ensuremath{\mathrm{MLPE}}\xspace} 
\newcommand{\deff}{\ensuremath{\mathrm{TEff}}\xspace} 
\newcommand{\ntokens}{t}

As our main evaluation metric, we report our models' perplexity ($\boldsymbol{\ppl}$) on the test set.
Further, we define three metrics that allow for easy comparison of monolingual and multilingual models. 
Let $\ntokens_\langa$ and $\ntokens_\langb$ be the number of tokens a multilingual model is trained on in languages $\lang_\langa$ and $\lang_\langb$, respectively.
We define monolingual token equivalence ($\boldsymbol{\monotok}$) as the number of tokens that would be required by a monolingual model, trained only in either language $\lang_\langa$ or $\lang_\langb$, to achieve the same perplexity as the multilingual model does in that language.
To determine $\monotok$, we fit a simple scaling law to predict perplexity from the number of training tokens (e.g., $\ntokens_\langa$) using the results from our trained monolingual models (see \cref{app:appl_scaling_laws} for details).
Analogously, we define monolingual $\ppl$ equivalence ($\boldsymbol{\mlpe}$) as the perplexity a monolingual model 
would reach when trained on the same number of $\lang_\langa$ tokens (i.e., $\ntokens_\langa$) as a given multilingual model.
Finally, we define token efficiency ($\boldsymbol{\deff}$) as the fraction between \monotok and the number of tokens used for multilingual training, e.g.,
$\deff_\langa = \frac{\monotok_\langa}{\ntokens_\langa}$.
Intuitively, if $\deff > 1$, performance improves due to multilinguality, while if $\deff < 1$, multilinguality hurts performance.\looseness=-1

\section{Cloned Languages}
\sectionvspace
\vspace{-2pt}
\label{sec:cloned_languages}

In this section, we examine the model's capability to generalise across perfectly equivalent \defn{cloned languages}. 
We create a cloned language by duplicating the language model's vocabulary; this allows us to encode each sequence in either the original language (using the original vocabulary) or in the cloned language (using the cloned vocabulary).
This experimental paradigm was originally proposed by \citet{wang2019cross} and \citet{dufter2020identifying}.\footnote{\citet{wang2019cross} perform duplication on the character IDs, i.e., before tokenisation, while \citet{dufter2020identifying} adopt an approach equivalent to ours.
Both of these works term $\lang_2$ a ``fake'' language. Since there is no distinction between $\lang_1$ and $\lang_2$, however, we call them cloned languages instead. 
Other related studies have investigated the effect of infinitely many cloned languages on LMs'
performance \citep{huang2023lexinvariant, chen2023improving}, or employed duplicated vocabularies at the token level to study their impact on LMs' memorisation or performance \citep{kharitonov2021bpe, schafer2024on}.\looseness=-1}
Formally, let $\langorig$ be an ``original'' language with a vocabulary of subword units $\vocab$; we denote each subword $\word \in \vocab$.
This language can be described by a probability distribution $p(\wordsorig)$, where $\wordsorig \in \vocab^{*}$.
We clone language $\langorig$ by creating multiple instantiations of it: $\lang_1$, $\lang_2$, \dots $\lang_N$. 
These languages have vocabularies $\vocab_n$, each of which has symbols that are equivalent to the original ones.\footnote{Unless otherwise noted, these vocabularies are defined as disjoint sets in our experiments, meaning that no anchor points exist across languages.}
Furthermore, these languages define probability distributions which are isometric to the original language.
If we denote equivalence as $\words_n \cloneequiv \wordsorig$ for $\words_n \in \vocab_n^*$ and $\wordsorig \in \vocab^*$, we have
$\words_n \cloneequiv \wordsorig \implies p(\words_n) = p(\wordsorig)$.\looseness=-1

\newcommand{\mathcomment}[1]{\text{\color{gray}{#1}}}

Given dataset $\dataorig = \{\wordsorig^{(k)}\}_{k=1}^K$ with $\wordsorig^{(k)} \sim p(\wordsorig)$, we can now create a multilingual dataset $\datamulti$ by independently mapping each sequence to one of the cloned languages: For each $\words^{(k)}$, we first sample a language $\lang^{(k)} \sim p(\lang)$ from a categorical distribution over languages, then we map the sequence to $\lang^{(k)}$ by encoding it using the corresponding vocabulary. We can write $\datamulti = \bigcup_{n=1}^N\dataset_{n}$ where
\vspace{-5pt}
\begin{equation*}
\dataset_{n} = \left\{\words_n^{(k)} \bigm\vert \words_n^{(k)} \cloneequiv \wordsorig^{(k)} \text{ and } \lang^{(k)} = \lang_n \right\}    
\vspace{-5pt}
\end{equation*}
denotes the subset in language $\lang_n$.

Importantly, cloned languages are perfectly equivalent, having the same syntax, semantics, and distribution.
They differ only in the symbols used to encode their vocabularies.
Any generalisation we observe in this setting should thus serve as an upper bound on the potential to generalise across non-identical natural languages.\footnote{
As for most of our experiments we consider cloned languages' alphabets to be disjoint, in practice our results only upper bound the cross-lingual generalisation of models with no anchor points (i.e., with disjoint vocabularies).\looseness=-1
}
In other words, if our model cannot generalise across cloned languages, we would have strong reason to believe it shouldn't generalise across distinct languages.
If we observe that a model can generalise across cloned languages, however, we may or may not observe the same to happen across non-cloned languages. We'll investigate the latter in Section \ref{sec:real_languages}.

\vspace{-2pt}
\subsection{Generalisation}
\subsectionvspace

Due to the equivalence of cloned languages, one may expect language models to easily generalise across them.
In that case, training a multilingual model on datasets $\dataset_1$ and $\dataset_2$ would lead to similar performance to training a monolingual model on the original dataset $\dataorig$ (note that $|\dataorig| = |\dataset_1| + |\dataset_2|$). 
We perform this experiment here, training either monolingual models on English ($\english$), or multilingual models on cloned English ($\english_1$ and $\english_2$), setting $p(\english_1) = p(\english_2) = 0.5$.
Perhaps surprisingly, when training in this balanced multilingual setting, language modelling performance is significantly worse than in the monolingual setting (see \cref{tab:synthetic_basic}, rows 2 \& 4).
In fact, one would obtain better performance training two monolingual models for half as many steps than training on this combined data.
Training data in one language seems to hurt performance in the other language instead of boosting it.
This indicates that the model is not able to generalise well across languages in this setting.\looseness=-1

\begin{takeaway}
The model does not generalise well across cloned languages given a \sfrac{50}{50} data split.\looseness=-1
\end{takeaway}

\newcommand{\tablepad}{}%

\newcolumntype{P}{>{\hspace{7pt}}c<{\hspace{7pt}}}
\newcolumntype{Q}{>{\hspace{7pt}}r<{\hspace{7pt}}}
\newcolumntype{S}{>{\hspace{0pt}}l<{\hspace{6pt}}}

\begin{table*}[t]
\centering
\resizebox{\textwidth}{!}{%
\begin{tabular} {SSQPPPPPPP}
\toprule
&& \multicolumn{4}{c}{\textbf{Training Data}} & \multicolumn{2}{c}{\textbf{PPL}} & \multicolumn{2}{c}{$\boldsymbol{\deff}$}\\ 
\cmidrule(lr){3-6}\cmidrule(lr){7-8}\cmidrule(lr){9-10}
\textbf{Run Type} & \textbf{Row} 
& \# Tokens & $p(\english_1)$   & $p(\english_2)$  & $p(\english_3), \dots, p(\english_{10})$ & $\english_1$   & $\english_2$ & $\english_1$   & $\english_2$  \\
\midrule
\multirow{3}{*}{Monolingual} &
1 & 1.2B & 100\% & 0\%  & 0\%                   & 21.9      & -      & 1 & -   \label{row:en12_100}\\ 
& 2 & 0.5 $\times$ 1.2B & 100\% & 0\%  & 0\%    & 25.3       & -      & 1 & -    \\ 
& 3 & 0.1 $\times$ 1.2B & 100\% & 0\%  & 0\%   & 48.4       & -     & 1 & -  \\ 
\midrule
\multirow{2}{*}{2 languages} &
4 & 1.2B & 50\% & 50\%  & 0\%        & 26.1  & 26.1   & 0.89 & 0.89      \label{row:en12_50_50} \\ 
& 5 & 1.2B & 90\% & 10\%  & 0\%        & 22.5  & 32.8  & 1.00 & 2.08        \\ 
\midrule
\multirow{2}{*}{10 languages} &
6 & 1.2B & 10\% & 10\% & 10\%, \dots, 10\% & 35.5  & 35.7    & 1.69 & 1.67      \\ 
& 7 & 1.2B & 50\% & $\frac{1}{18}$ & $\frac{1}{18}, \dots, \frac{1}{18}$ & 24.6  & 33.4 & 1.15 & 3.56        \\ 
\midrule
\multirow{2}{*}{Schedule} &
8 & 1.2B &\stepDown{100\%}{0\%} & \stepUp{0\%}{100\%}  & 0\% & $>$1B  & 31.4  & - & 0.47  \\ 
& 9 & 1.2B & \stepDown{90\%}{10\%} & \stepUp{10\%}{90\%}  & 0\% & 26.5  & 24.4 & 0.83 & 1.18 \\
\midrule
\multirow{2}{*}{2x data} &
10 & 2 $\times$ 1.2B & 50\% & 50\%   & 0\% & 23.3  & 23.3    & 0.73 & 0.73      \\ 
& 11 & 2 $\times$ 1.2B & \stepDown{90\%}{10\%} & \stepUp{10\%}{90\%} & 0\% & 22.8  & 20.4 & 0.83 & 1.60 \\
\midrule
\multirow{2}{*}{3x data} &
12 & 3 $\times$ 1.2B & 50\% & 50\%  & 0\% & 22.2  & 22.2  & 0.64 & 0.64     \\ 
& 13 & \tablepad 3 $\times$ 1.2B \tablepad & \tablepad \stepDown{90\%}{10\%} \tablepad & \tablepad \stepUp{10\%}{90\%}  \tablepad & 0\% & 21.5  & 19.3 & 0.77 & 1.63 \\
\bottomrule
\end{tabular}%
}
\caption{Performance of language models trained on different compositions of $\english_1$ and $\english_2$. \stepDown{a\%}{b\%} indicates an immediate decrease from a\% down to b\% halfway during training. Analogously, \stepUp{a\%}{b\%} indicates an immediate increase.\looseness=-1
}
\label{tab:synthetic_basic}
\vspace{-10pt}
\end{table*}

\begin{figure}[t]
    \centering
    \includegraphics[width=0.95\linewidth]{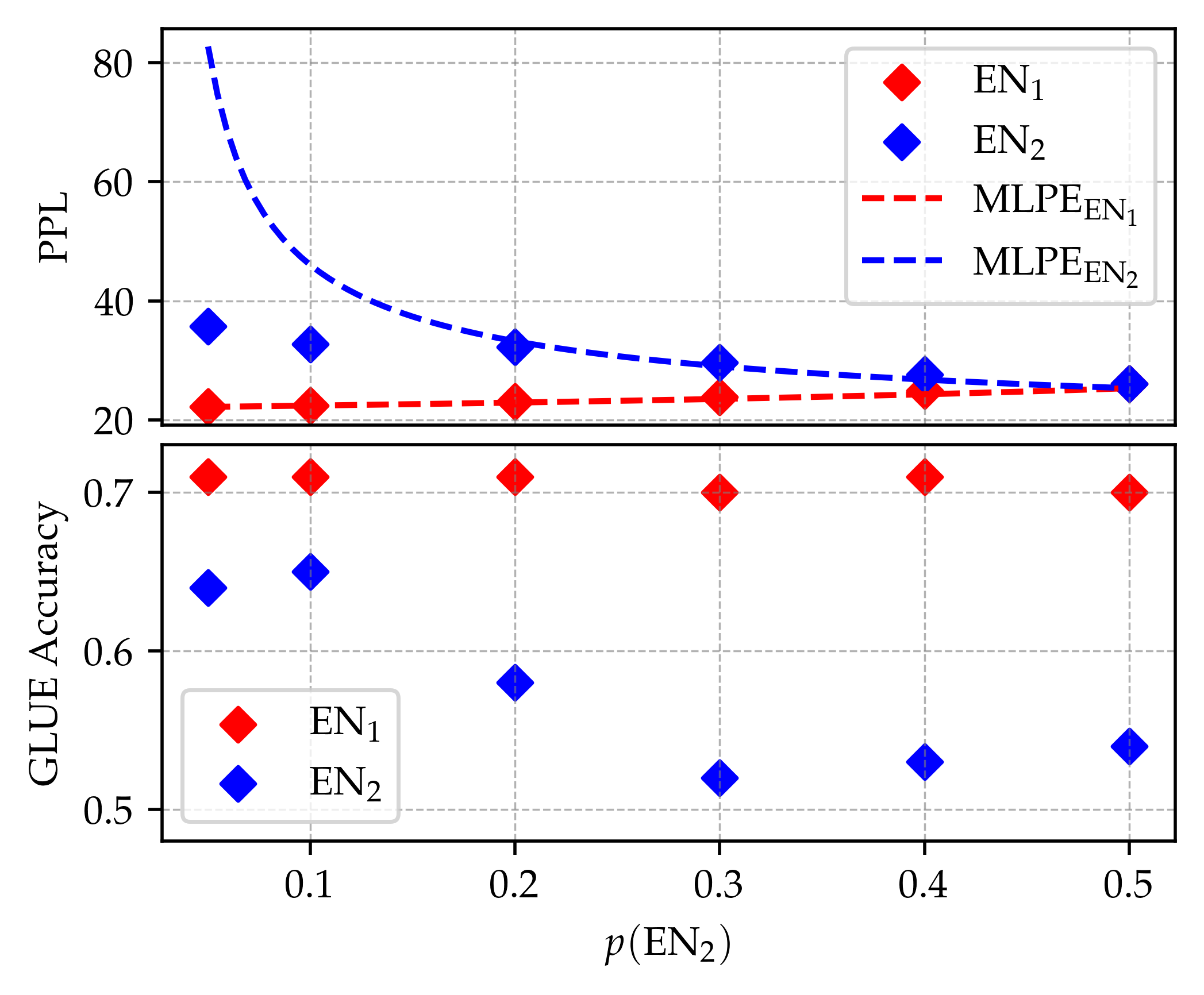}
    \caption{
    LM performance by imbalance ratio.
    (top) LM perplexity.
    (bottom) LM accuracy on GLUE; models were fine-tuned in $\english_1$ and evaluated on either $\english_1$ and $\english_2$.\looseness=-1
    }
    \label{fig:language_inbalance_main}
\vspace{-10pt}
\end{figure}

\subsection{Language Imbalance}
\subsectionvspace
\label{sec:e12_imbalance}

How does the balance of the languages' data affect generalisation performance?
Will the multilingual model still underperform its monolingual equivalents when trained on an uneven language distribution?
When varying the ratio of $\english_1$ to $\english_2$ data shown during training (while keeping the total number of training steps constant), we observe that the rarer ``lower resource'' language, here always $\english_2$, benefits from the presence of a dominant ``main language''.
\cref{fig:language_inbalance_main} (left) shows that, under higher imbalance, the model's performance on $\english_2$ becomes much better than that of a monolingual model trained on the same amount of $\english_2$ data. 
For example, when training in the \sfrac{90}{10} regime, we obtain a $\deff_{\english_2}$ of over 2 (see \cref{tab:synthetic_basic}, row 5). 
Do these improvements translate to better cross-lingual generalisation on downstream tasks as well?
We test this by fine-tuning models on the GLUE benchmark \citep{wang2019glue} in $\english_1$ only, and evaluating them on $\english_1$ and $\english_2$.
We observe that models trained under higher language imbalance indeed have significantly better $\english_2$ zero-shot performance (see \cref{fig:language_inbalance_main} right).
Together, these results suggest that cross-lingual generalisation is occurring.\looseness=-1

Is this generalisation attained due to the model's internal computations being shared across languages?
To answer this question, we analyse how language imbalance affects the cross-lingual alignment of our models' representations.
Looking at the cosine similarity of equivalent subwords 
$\word_1 \cloneequiv \word_2$ in 
$\english_1$ and $\english_2$, we find that similarity steadily increases with higher imbalance: in the \sfrac{50}{50} setting, embeddings are not aligned (exhibiting an average cosine similarity of 0.02), while, e.g., in the \sfrac{90}{10} setting, equivalent subwords are much more aligned, showing a similarity of 0.28 (details in \cref{app:alignment_en12}).
Comparing the cosine similarity of hidden states when the LM is given equivalent sequences $\words_1 \cloneequiv \words_2$, we also observe stronger alignment for a model trained in the imbalanced \sfrac{90}{10} regime, compared to the \sfrac{50}{50} counterpart (see \cref{app:hidden_sim}).
Interestingly, the cosine similarity between gradients is also higher in the imbalanced setting: when processing equivalent sequences, the gradients with respect to $\words_1$ or $\words_2$ have an average cosine similarity of 0.53 for the model trained in the \sfrac{90}{10} setting, compared to 0.07 in the \sfrac{50}{50} setting (see full plots of similarities per model component in \cref{app:gradient_similarity}).
This suggests that the gradient updates with respect to one language may benefit the optimisation process of that language's cloned counterpart more when training under higher imbalance.

\begin{takeaway}
    Language imbalance improves generalisation and leads to representations which are more aligned across cloned languages.
\end{takeaway}

\subsection{Many Languages}
\subsectionvspace

How does this trend transfer to settings with more than two languages? 
In such cases, sharing circuits
across languages might be even more crucial due to the model's limited capacity.
Instead of cloning the language only once, we now clone it nine times, obtaining in total 10 languages.
In \cref{tab:synthetic_basic} (rows 6 \& 7), we report the performance when sampling the languages in a balanced way and when having a much stronger main language.

Interestingly, when sampling uniformly, we obtain $\deff \approx 1.7$; performance is thus better than with a monolingual model trained on an equivalent amount of monolingual data (compare rows 6 \& 3). 
This differs from our observations for the bilingual setting, where uniform language sampling performed worse than the equivalent monolingual models. 
Presumably, modelling this many languages effectively with limited model capacity may lead the model to share its circuits, improving cross-lingual generalisation \citep{dufter2020identifying}.
The limit of infinite languages (in which a model never observes the same language more than once) was analysed by \citet{huang2023lexinvariant}; interestingly, LMs still seem to learn the language, to some extent, even in that setting.

In the imbalanced setting where we sample a stronger ``main language'' 50\% of the time, we observe even stronger performance on all languages. Despite the model seeing only roughly 67M tokens in each of the rarer languages (1/18 of all steps), it achieves \textbf{better} performance in these languages than in the uniform setting with 120M tokens (1/10 of all steps) per language. In fact, on the rarer languages, the model achieves $\deff \approx 3.6$, matching the performance of a monolingual model trained on 240M tokens.\looseness=-1

\begin{takeaway}
    When training on many cloned languages, sampling a main language disproportionately improves generalisation.
\end{takeaway}

\begin{figure}[t]
    \centering
    \includegraphics[width=\linewidth]{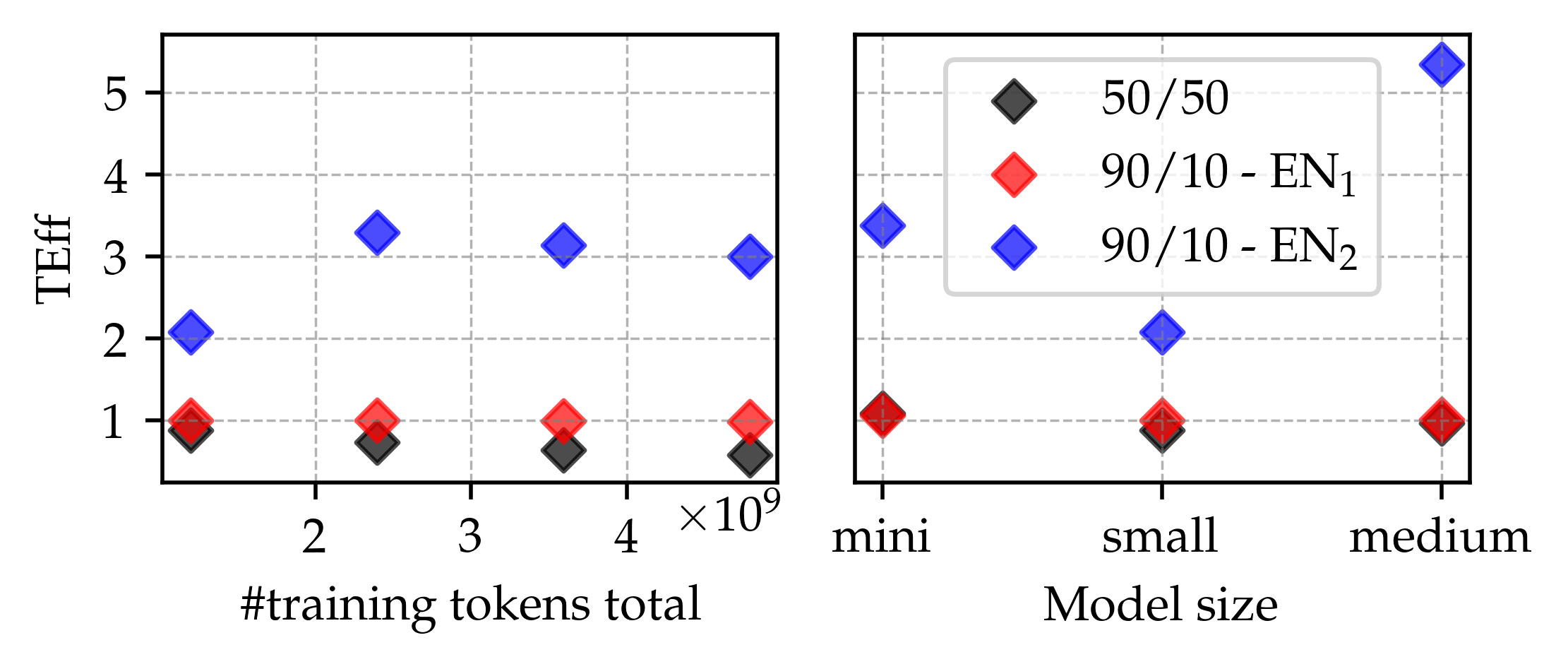}
    \vspace{-10pt}
    \caption{$\deff$ as we train LMs with (left) more data, or (right) larger architectures. mini, small and medium denote GPT sizes in Languini \citep{stanic2023languini}, with 11M, 85M, and 303M non-embedding parameters.\looseness=-1}
    \label{fig:more_data_and_larger_models}
    \vspace{-10pt}
\end{figure}

\subsection{Effect of scaling}
\subsectionvspace

Model and data size are crucial factors for the performance of LMs. 
Here, we investigate how the previously identified trends are affected by scaling the model architecture or training data. 
\cref{fig:more_data_and_larger_models} (left) shows that the effect of imbalance on cross-lingual generalisation appears to increase when we train on twice as much data (2.4B tokens instead of 1.2B), reaching $\deff > 3$; 
this corresponds to a ``chinchilla-optimal'' setup for our GPT small model \citep{hoffmann2022chinchilla}.
At the same time, the $\deff$ of the \sfrac{50}{50} setting seems to be decreasing under prolonged training. This might be caused by the heightened importance of model capacity under longer training, which may have a stronger impact on performance when representations are less aligned across languages. 
Overall, the disparity in effectiveness between the imbalanced and balanced settings grows with longer training.
Remarkably, when training for 4.8B tokens, the \sfrac{90}{10} setting yields better performance in both languages, compared to the \sfrac{50}{50} setting.\looseness=-1

When decreasing the model size, we also observe higher performance benefits in the imbalanced setting
(see \cref{fig:more_data_and_larger_models} right), potentially due to the capacity argument described above.
Interestingly, however, the effect of imbalance appears to be significantly stronger for larger models as well.
When training a larger model with around 300M parameters \citep[GPT medium in Languini;][]{stanic2023languini}, in the \sfrac{90}{10} setting, we achieve
better performance on both languages than under the \sfrac{50}{50} split. 
This might be because larger models generally exhibit better generalisation ability than smaller ones \citep{brown2020gpt3}.\looseness=-1

\begin{takeaway}
Longer training and larger models lead to stronger performance benefits due to language imbalance.
\end{takeaway}

\subsection{Language Sampling Schedule}
\subsectionvspace

Knowing that language imbalance boosts generalisation, how can we use this insight to train better models? 
Is there a way to leverage our insights in order to improve performance on two languages, even with the same training data?
In \cref{tab:synthetic_basic} (rows 8, 9, 11, and 13), we report results when training with a language sampling schedule that ensures a language imbalance throughout all of training, but which still leads to an overall \sfrac{50}{50} split between $\english_1$ and $\english_2$ data seen by the model.
We sample $\english_1$ with a higher probability during the first half of training.
Then, we sample $\english_2$ more often to achieve a marginal split of \sfrac{50}{50}.\looseness=-1

When showing exclusively $\english_1$ at first, and then showing only $\english_2$ (\sfrac{\stepDown{100}{0\,\,}}{\stepUp{\,\,0}{100}}; row 8), we observe bad overall performance. 
By the end of training, perplexity on $\english_1$ is very high, presumably due to catastrophic forgetting \citep{mccloskey1989catastrophic,french1999catastrophic}.
Further, $\english_2$ does not seem to benefit from the $\english_1$ data, achieving very low performance, which might be due to the lower learning rate in the second half of training. \footnote{\citet{chen2023improving} find that an equivalent setting can still be beneficial when using many more languages: they periodically reinitialise the learned embeddings (which is equivalent to switching to a new cloned language) and obtain models that are better adaptable to new languages.}

On the other hand, if we avoid catastrophic forgetting, making sure that the model encounters at least some samples of both languages at every point in training, via a 
\sfrac{\stepDown{90}{10}}{\stepUp{10}{90}} split
(first sampling languages with ratio \sfrac{90}{10}, and then switching to \sfrac{10}{90} after half of training), we can mitigate these issues.
On our standard training set (1.2B tokens, row 9), we observe almost equivalent performance to uniform language sampling on $\english_1$, but significantly improved performance on $\english_2$.
Under longer training, these benefits become more pronounced: this language schedule 
improves performance on both languages compared to the simple \sfrac{50}{50} setting (compare row 10 vs 11 and row 12 vs 13).

\begin{takeaway}
    Compared to uniform language sampling, an imbalanced ratio throughout training can lead to better results on all languages,  even if the overall language split remains balanced.\looseness=-1
\end{takeaway}

\vspace{-5pt}
\section{Real Languages}
\sectionvspace
\label{sec:real_languages}
\vspace{-2pt}

To verify whether the insights from our cloned-language experiments hold in a more natural setting, we now run experiments with multilingual models on English ($\english$) and French ($\french$).\looseness=-1

\subsection{Generalisation}
\subsectionvspace

In the cloned setting, we observed no significant generalisation
when training on a balanced language mix (i.e., $\deff < 1$, representations were unaligned, and zero-shot GLUE accuracy on $\english_2$ was bad). 
Similarly, when sampling $\english$ and $\french$ data uniformly, we also obtain $\deff < 1$.
A multilingual model's performance is thus worse than a monolingual model trained only in the same $\english$ or $\french$ data (see \cref{tab:enfr_basic}, row 7).
Notably, prior work has identified anchors (shared vocabulary items across languages) help generalisation \citep{dufter2020identifying, pires2019multilingual, wu2019beto}.
We thus experiment with similarly merging vocabulary items shared between $\english$ and $\french$, and confirm this helps performance (compare \cref{tab:enfr_basic}, row 7 vs 11).
We run more experiments analysing the impact of anchor points in both cloned and real languages, see \cref{app:full_anchors}. 
Note that, with an anchored vocabulary, generalisation across $\english$ and $\french$ is not necessarily upper bounded by our results on disjoint cloned languages. 
In fact, in the \sfrac{50}{50} setting, we observe a marginally higher $\deff$ for $\english$--$\french$ models with an anchored vocabulary than for $\english_1$--$\english_2$ models where we used disjoint vocabularies (compare \cref{tab:synthetic_basic} row 4 and \cref{tab:enfr_basic} row 11). \looseness-1

\tiago{Do we want a takeaway here to have one per section? Kind of unnecessary, but nice for consistency? ...\response{anton} will try to write one if I there's time}

\begin{table*}[t]
\centering
\resizebox{\textwidth}{!}{%
\begin{tabular}{SSQPPPPPP}
\toprule
&& \multicolumn{3}{c}{\textbf{Training Data}} & \multicolumn{2}{c}{\textbf{PPL}} & \multicolumn{2}{c}{$\boldsymbol{\deff}$}\\ 
\cmidrule(lr){3-5}\cmidrule(lr){6-7}\cmidrule(lr){8-9}
\textbf{Run Type} & \textbf{Row} 
& \# Tokens & $p(\english)$   & $p(\french)$  & $\english$   & $\french$ & $\english$   & $\french$  \\
\midrule
\multirow{6}{*}{Monolingual } &
1 & 1.2B & 100\% & 0\%                    & 21.9      & -      & 1 & -  \\ 
& 2 & 0.5 $\times$ 1.2B & 100\% & 0\%     & 25.3       & -      & 1 & -    \\ 
& 3 & 0.1 $\times$ 1.2B & 100\% & 0\%   & 48.4       & -     & 1 & -  \\ 
& 4 &1.2B & 0\% & 100\%                    & - & 16.0      & -      & 1    \\ 
& 5 & 0.5 $\times$ 1.2B & 0\% & 100\%     & -       & 18.4      & - & 1    \\ 
& 6 & 0.1 $\times$ 1.2B & 0\% & 100\%   & - & 34.1       & -     & 1   \\ 
\midrule
\multirow{4}{*}{\parbox{2.5cm}{Multilingual \\ disjoint vocabs}} &
7 & 1.2B & 50\% & 50\%       & 26.4  & 19.4   & 0.85 & 0.82     \\ 
& 8 & 1.2B & 90\% & 10\%        & 22.5  & 31.9  & 1.00 & 1.05         \\ 
& 9 & 1.2B & 10\% & 90\%        & 43.5  & 16.4  & 1.10 & 0.97         \\ 
& 10 & 1.2B & \stepDown{90\%}{10\%} & \stepUp{10\%}{90\%}     & 29.1  & 20.5   & 0.60 & 0.66     \\ 
\midrule
\multirow{6}{*}{\parbox{2.6cm}{Multilingual \\ anchored vocabs}} &
11 & 1.2B & 50\% & 50\%       & 26.0  & 19.0   & 0.91 & 0.88     \\ 
& 12 & 1.2B & 90\% & 10\%        & 22.5  & 29.0  & 1.00 & 1.27         \\ 
& 13 & 1.2B & 10\% & 90\%        & 39.5  & 16.5  & 1.33 & 0.96         \\ 
& 14 & 1.2B & \stepDown{90\%}{10\%} & \stepUp{10\%}{90\%}     & 28.9 & 19.3 & 0.61 & 0.83     \\ 
& 15 & 1.2B & 90\%\downSymbol10\% \upSymbol 50\% \staySymbol 50\% & 10\%\upSymbol90\% \downSymbol 50\% \staySymbol 50\%      & 26.4  & 18.5   & 0.85 & 1.00     \\ 
& 16 & 1.2B & 95\%\downSymbol35\% \staySymbol 35\% \staySymbol 35\% & 5\%\upSymbol65\% \staySymbol 65\% \staySymbol 65\%      & 26.3  & 18.7   & 0.86 & 0.95     \\ 
\cmidrule(lr){2-9}
\multirow{2}*{\parbox{2.5cm}{2x data}} &
17 & 2 $\times$ 1.2B & 50\% & 50\%       & 23.0  & 16.9   & 0.79 & 0.76     \\ 
& 18 & 2 $\times$ 1.2B & \stepDown{90\%}{10\%} & \stepUp{10\%}{90\%}     & 26.1  & 17.1   & 0.44 & 0.70     \\ 
\cmidrule(lr){2-9}
\multirow{2}*{\parbox{2.5cm}{3x data}} &
19 & 3 $\times$ 1.2B & 50\% & 50\%       & 21.8  & 16.0   & 0.70 & 0.67     \\ 
& 20 & 3 $\times$ 1.2B & \stepDown{90\%}{10\%} & \stepUp{10\%}{90\%}     & 25.1  & 16.2   & 0.35 & 0.63     \\ 
\bottomrule
\end{tabular}%
}
\caption{Performance of language models trained on different compositions of $\english$ and $\french$.
a\%\staySymbol b\%\staySymbol c\% \staySymbol d\% indicates a four stage language schedule, switching immediately between, e.g., c\% and d\% after 75\% of training. %
\looseness=-1}
\label{tab:enfr_basic}
\vspace{-10pt}
\end{table*}

\subsection{Language Imbalance} \label{sec:reallangs_imbalance}
\subsectionvspace

\begin{figure}[t]
    \centering
    \includegraphics[width=0.95\linewidth]{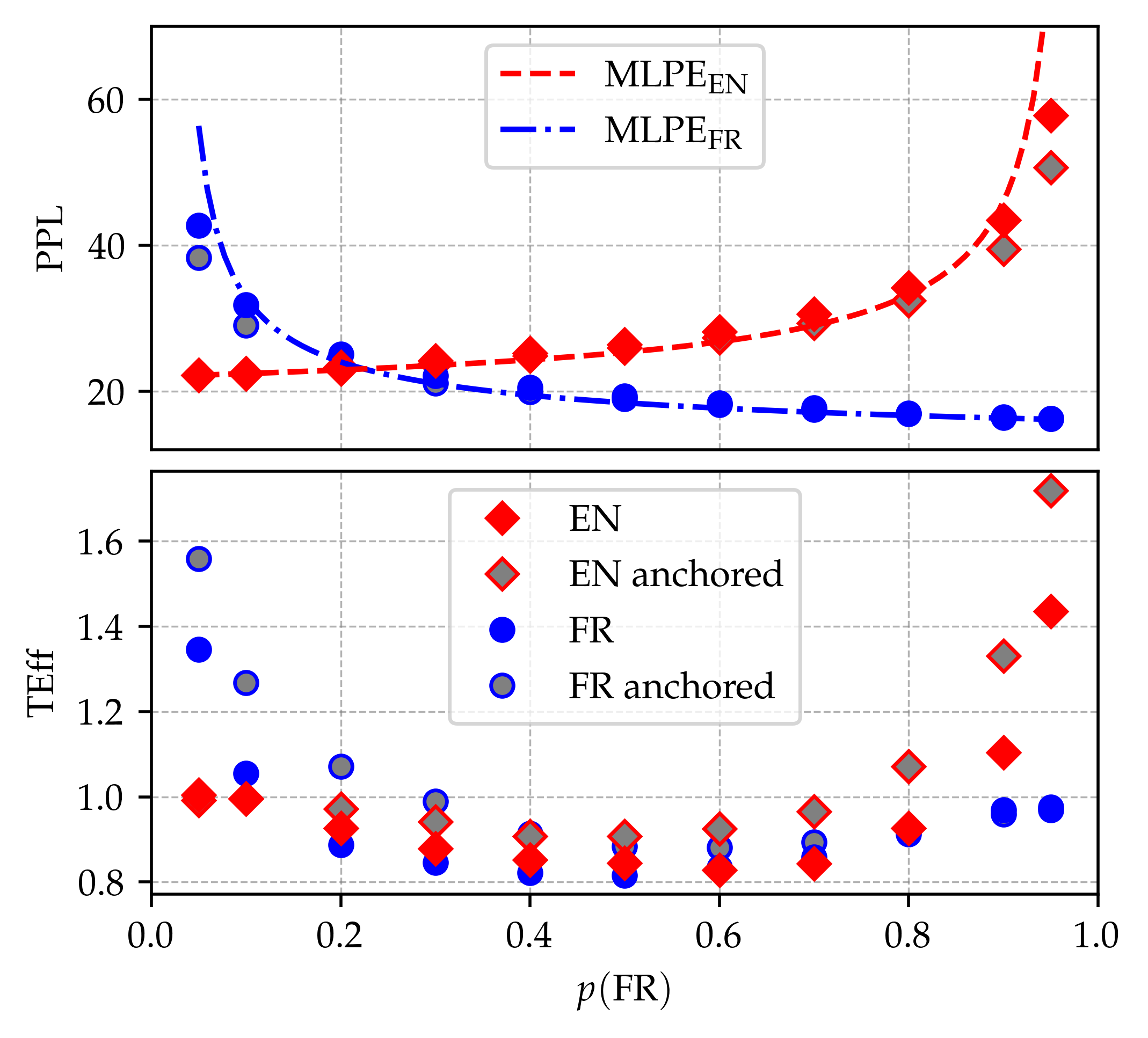}
    \caption{LM performance on $\english$ and $\french$ by imbalance ratio.}
    \label{fig:ppl_dataeff_en_french_ratio}
\vspace{-10pt}
\end{figure}

Analogous to the cloned setting, we observe that an imbalanced $\sfrac{\english}{\french}$ ratio leads to improved performance ($\deff > 1$), on the rarer language (see \cref{tab:enfr_basic}, rows 7-9 \& 11-13). 
This is the case for both, a \sfrac{90}{10} and a \sfrac{10}{90} $\sfrac{\english}{\french}$ ratio.
\cref{fig:ppl_dataeff_en_french_ratio} shows $\ppl$ and $\deff$ in $\english$ and $\french$ as a function of the language imbalance.
We observe that large imbalances generally seem to yield $\deff > 1$; the worst $\deff$ is reached with a balanced $\sfrac{\english}{\french}$ ratio.
These trends are in line with our findings in the cloned setting. 
However, especially with disjoint vocabularies, the observed performance benefits due to generalisation are less significant. 
Presumably, this is due to $\english$ and $\french$ not being equivalent and thus generally allowing less generalisation.\looseness=-1

Does imbalance again improve generalisation due to a better alignment of the model's representations in the two languages?
As in the cloned language setting, we investigate the cosine similarity between the models' hidden states when processing ``equivalent'' sequences in the two languages.
For real languages, however, we do not have access to perfectly equivalent sequences. 
Instead, we mimick this scenario
using parallel translated sequences in the two languages, which should contain roughly similar properties.
Differently from the cloned language setting, we do not observe higher hidden state similarities for models trained on imbalanced data (see \cref{app:hidden_sim}).
Further, we find that gradient similarities barely differ across balanced and imbalanced settings when using disjoint vocabularies. For the anchored vocabulary they are even marginally higher in the balanced setting (see \cref{app:gradient_similarity}).
We thus do not find evidence that the improved $\deff$ in the imbalanced setting is caused by a stronger alignment of model updates across languages
in this setting. 
A possible reason for this discrepancy could be that, at the scales of our experiments, LMs tend to rely on language specific surface-level features (which are shared by cloned languages, but not by distinct real languages) and show less understanding of complex semantics which might be more generalisable. Future research might thus consider investigating these trends at larger scales.

\begin{takeaway}
    Imbalanced multilinguality boosts the performance of real low-resource languages. However, this effect is weaker here than for cloned languages. Further, for real languages, we do not find evidence of language imbalance leading to representations which are more cross-lingually aligned.\looseness=-1
\end{takeaway}

\vspace{-2pt}
\subsection{Effect of Scaling}
\label{sec:enfr_scaling}
\subsectionvspace

In the cloned setting, we observed that prolonging training significantly decreased $\deff$ in the \sfrac{50}{50} setting.
We hypothesised that this might be caused by a stronger influence of the limited model capacity with longer training, and poor sharing of representations between languages.
As $\english$ and $\french$ are distinct languages that require at least some language specific representations, we might expect this trend to be even more pronounced for these languages.
However, compared to the cloned setting, prolonging training leads to a smaller decline in $\deff$ in the \sfrac{50}{50} setting here.
Presumably, the anchored vocabulary allows for better generalisation compared to the cloned setting, despite the languages being distinct.

\begin{figure}[t]
    \centering
    \includegraphics[width=\linewidth]{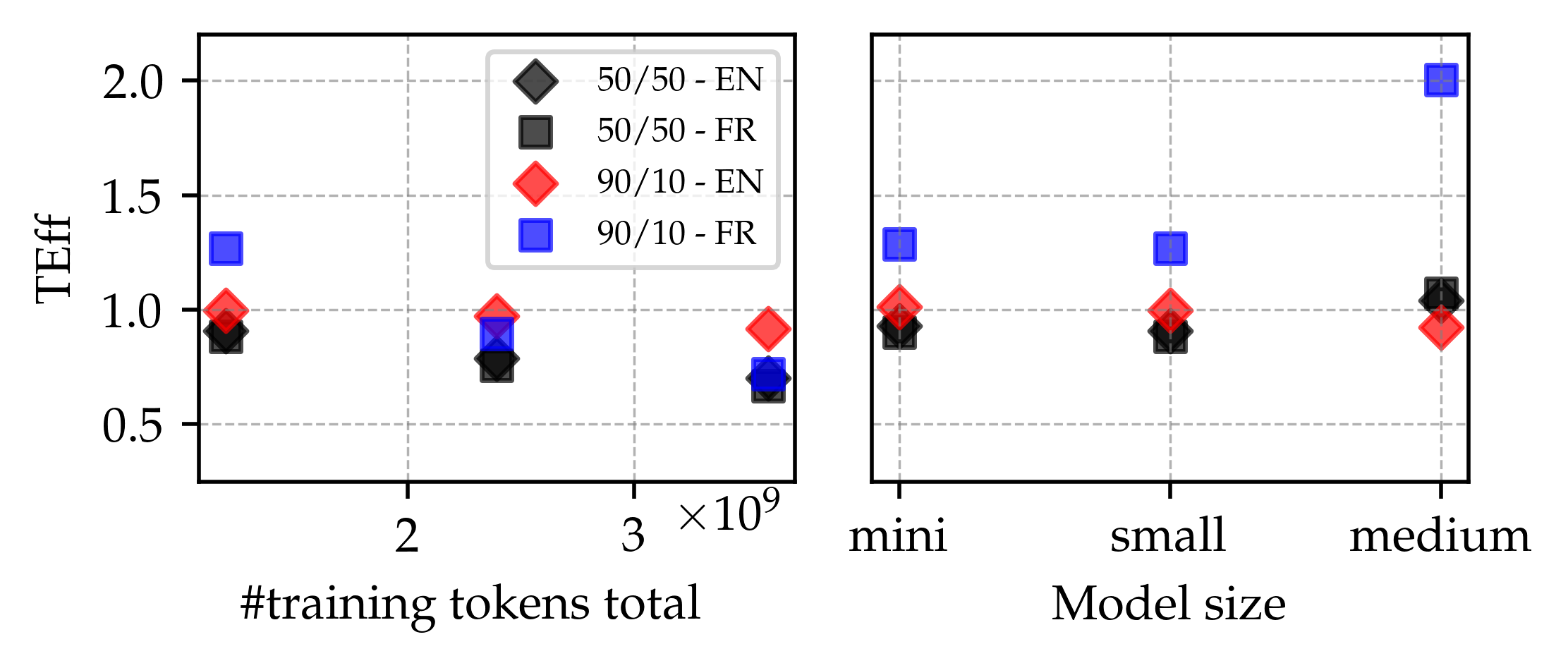}
    \vspace{-19pt}
    \caption{$\deff$ of models on $\english$ and $\french$ with anchored vocab as we train them with (left) more data, or (right) larger architectures. 
    }
    \label{fig:enfr_scaling}
    \vspace{-8pt}
\end{figure}

Further, unlike in the cloned setting,
longer training significantly decreases the $\deff$ of the lower-resource language in the imbalanced setting here
(see \cref{fig:enfr_scaling}). 
In fact, the \sfrac{90}{10} $\deff$ even falls below 1, approaching the $\deff$ of the \sfrac{50}{50} setting.
This suggests that language imbalance might not improve generalisation across distinct real languages.
Still, when scaling up the model, we observe an increase of almost 2x in the $\deff$ of the lower-resource language (see \cref{fig:enfr_scaling}). 
This is in line with our cloned languages observations, although the effect is weaker.\looseness=-1

\begin{takeaway}
Performance benefits for real low-resource languages tend to decrease or vanish with longer training. Larger models, however, appear to yield higher performance benefits in both the balanced and imbalanced setting. 
\end{takeaway}

\vspace{-6pt}
\subsection{Language Sampling Schedule}
\subsectionvspace

For equivalent cloned languages, we found that an imbalanced language sampling schedule can lead to improvements upon simple uniform sampling.
If this held for real languages as well, it could have important practical implications for future multilingual LM training.
However, whereas a \sfrac{\stepDown{90}{10}}{\stepUp{10}{90}} schedule yielded strong performance on cloned languages, matching or outperforming the \sfrac{50}{50} setting, this is not the case for $\english$ and $\french$ (see \cref{tab:enfr_basic}, row 10 vs 7 and row 14 vs 11). 
Furthermore, in line with the observations above, longer training does not make this schedule more effective, but instead increases its gap to the performance of the \sfrac{50}{50} setting (see rows 17-20).

The discrepancy between these results and the ones in cloned languages might be explained by the reduced effect of imbalance on the generalisation and representation alignment in real languages.
The schedules may be enough to force LMs to share circuits across cloned languages, but not across real ones.
To investigate if this negative result was a particular property of our chosen schedule, we explore other more complex scheduling options.
\footnote{Future research might design these more carefully, also analysing the interplay of language- and learning rate schedule\looseness=-1}
In general, none of the tested schedules appears to outperform the \sfrac{50}{50} setting (see rows 15, 16) on both languages.
However, more complex 4-stage schedules, can obtain better performance on one language while incurring a slight performance drop in the other.
Intriguingly, this allows trading off the performance of different languages without altering the training data.

\begin{takeaway}
    For real languages, we do not find improvements on all languages due to the tested language schedules. However, they allow for trading off performance in different languages.
\end{takeaway}

\vspace{-6pt}
\section{Conclusion}
\sectionvspace

We ran experiments to measure cross-lingual generalisation in both a controlled setting with cloned English languages, as well as with English and French.
In both settings, we find that, without vocabulary overlap, our models do not show strong cross-lingual generalisation when trained on a balanced language set.
However, when training on an imbalanced mix of languages, we observe increased performance compared to monolingual settings.
For cloned languages, we find that this can be explained by a higher alignment of the model's representations across languages, which indicates circuit reuse and improved cross-lingual generalisation. 
Yet, at the scales of our experiments, such a correlation is less evident in real languages. 
While our findings allow us to design an imbalanced language schedule that yields improved performance in the cloned setting, further research is required to extend these improvements to real-world settings.\looseness=-1

\section*{Limitations}
There are several limitations of our work, many of which present opportunities for future research.

\paragraph{Data and model size.} 
While we conduct experiments with varying data (up to 4.8B tokens) and model size (up to 336M parameters), it is uncertain whether the identified trends also apply at the scale of modern large language models. Additionally, for more capable models, cross-lingual generalisation might be relevant in different aspects, with, e.g., semantics playing a larger role. As the semantic content communicated in different languages might be easily transferable, this might impact generalisation dynamics.

\paragraph{Languages.} 
We only run experiments on English and French, two Indo-European languages. 
Further work could consider more languages and investigate the impact of language similarity in results more broadly.

\paragraph{Model architecture.} 
We run most of our experiments on a Transformer decoder (we also measure embedding alignment for simpler Word2Vec models). 
Future research could analyse the effects of architecture in more depth to better understand the drivers of representation alignment. \citet{conneau2020emerging}, e.g.,  find that shared parameters in the top layers lead to better cross-lingual transfer. In our Word2Vec experiments, we do not observe improvements in representation alignment due to language imbalance (see \cref{fig:cossim_synth_ratio}), presumably due to no parameters being shared between the two languages. Would this change when adding a shared layer to the Word2Vec model?

\paragraph{Downstream performance.} 
In our evaluation we mainly rely on perplexity as a metric, with a single experiment on GLUE accuracy. 
It might be insightful to analyze effects on downstream task performance more broadly.

\paragraph{Quantifying generalisation.}
In this work, we mainly measure cross-lingual generalisation by comparing the performance of multilingual models with that of monolingual models trained on the same amount of data in the given language. 
If a multilingual model on languages $\lang_\langa$ and $\lang_\langb$ requires fewer $\lang_\langa$ tokens to reach a given perplexity on $\lang_\langa$ than a monolingual model, we speak of cross-lingual generalisation, knowing that performance on $\lang_\langa$ must have been boosted by data in language $\lang_\langb$. Future work could formalise this measure and aim to model/quantify the relationship between the number of training tokens seen in a language $\lang_\langb$ and performance in another language $\lang_\langa$, depending on model size, language imbalance, language similarity, anchor points, and other factors. An accurate model of these relationships could be of substantial practical value.

\Anton{What about concept bias? what about grammar etc. in the lower resource language?}

\bibliography{custom}

\begin{thebibliography}{47}
\expandafter\ifx\csname natexlab\endcsname\relax\def\natexlab#1{#1}\fi

\bibitem[{Ahia et~al.(2023)Ahia, Kumar, Gonen, Kasai, Mortensen, Smith, and
  Tsvetkov}]{ahia-etal-2023-languages}
Orevaoghene Ahia, Sachin Kumar, Hila Gonen, Jungo Kasai, David Mortensen, Noah
  Smith, and Yulia Tsvetkov. 2023.
\newblock \href {https://aclanthology.org/2023.emnlp-main.614} {Do all
  languages cost the same? {T}okenization in the era of commercial language
  models}.
\newblock In \emph{Proceedings of the 2023 Conference on Empirical Methods in
  Natural Language Processing}, pages 9904--9923, Singapore. Association for
  Computational Linguistics.

\bibitem[{Alabi et~al.(2024)Alabi, Mosbach, Eyal, Klakow, and
  Geva}]{alabi2024hidden}
Jesujoba~O. Alabi, Marius Mosbach, Matan Eyal, Dietrich Klakow, and Mor Geva.
  2024.
\newblock \href {https://arxiv.org/abs/2402.13137} {The hidden space of
  transformer language adapters}.
\newblock \emph{arXiv preprint arXiv:2402.13137}.

\bibitem[{Blevins and Zettlemoyer(2022)}]{blevins2022language}
Terra Blevins and Luke Zettlemoyer. 2022.
\newblock \href {https://arxiv.org/abs/2204.08110.pdf} {Language contamination
  helps explains the cross-lingual capabilities of english pretrained models}.
\newblock In \emph{Proceedings of the 2022 Conference on Empirical Methods in
  Natural Language Processing}, pages 3563--3574.

\bibitem[{Briakou et~al.(2023)Briakou, Cherry, and
  Foster}]{briakou2023searching}
Eleftheria Briakou, Colin Cherry, and George Foster. 2023.
\newblock \href {https://arxiv.org/abs/2305.10266.pdf} {Searching for needles
  in a haystack: On the role of incidental bilingualism in {PaLM}’s
  translation capability}.
\newblock In \emph{Proceedings of the 61st Annual Meeting of the Association
  for Computational Linguistics (Volume 1: Long Papers)}, pages 9432--9452.

\bibitem[{Brown et~al.(2020)Brown, Mann, Ryder, Subbiah, Kaplan, Dhariwal,
  Neelakantan, Shyam, Sastry, Askell, Agarwal, Herbert-Voss, Krueger, Henighan,
  Child, Ramesh, Ziegler, Wu, Winter, Hesse, Chen, Sigler, Litwin, Gray, Chess,
  Clark, Berner, McCandlish, Radford, Sutskever, and Amodei}]{brown2020gpt3}
Tom Brown, Benjamin Mann, Nick Ryder, Melanie Subbiah, Jared~D Kaplan, Prafulla
  Dhariwal, Arvind Neelakantan, Pranav Shyam, Girish Sastry, Amanda Askell,
  Sandhini Agarwal, Ariel Herbert-Voss, Gretchen Krueger, Tom Henighan, Rewon
  Child, Aditya Ramesh, Daniel Ziegler, Jeffrey Wu, Clemens Winter, Chris
  Hesse, Mark Chen, Eric Sigler, Mateusz Litwin, Scott Gray, Benjamin Chess,
  Jack Clark, Christopher Berner, Sam McCandlish, Alec Radford, Ilya Sutskever,
  and Dario Amodei. 2020.
\newblock \href
  {https://proceedings.neurips.cc/paper_files/paper/2020/file/1457c0d6bfcb4967418bfb8ac142f64a-Paper.pdf}
  {Language models are few-shot learners}.
\newblock In \emph{Advances in Neural Information Processing Systems},
  volume~33, pages 1877--1901. Curran Associates, Inc.

\bibitem[{Cammarata et~al.(2020)Cammarata, Carter, Goh, Olah, Petrov, Schubert,
  Voss, Egan, and Lim}]{cammarata2020thread}
Nick Cammarata, Shan Carter, Gabriel Goh, Chris Olah, Michael Petrov, Ludwig
  Schubert, Chelsea Voss, Ben Egan, and Swee~Kiat Lim. 2020.
\newblock \href {https://distill.pub/2020/circuits} {Thread: Circuits}.
\newblock \emph{Distill}.

\bibitem[{Chang et~al.(2023)Chang, Arnett, Tu, and Bergen}]{chang2024when}
Tyler~A. Chang, Catherine Arnett, Zhuowen Tu, and Benjamin~K Bergen. 2023.
\newblock \href {https://arxiv.org/pdf/2311.09205.pdf} {When is multilinguality
  a curse? language modeling for 250 high-and low-resource languages}.
\newblock \emph{arXiv preprint arXiv:2311.09205}.

\bibitem[{Chen et~al.(2023)Chen, Marchisio, Raileanu, Adelani, Saito~Stenetorp,
  Riedel, and Artetxe}]{chen2023improving}
Yihong Chen, Kelly Marchisio, Roberta Raileanu, David Adelani, Pontus Lars~Erik
  Saito~Stenetorp, Sebastian Riedel, and Mikel Artetxe. 2023.
\newblock \href {https://arxiv.org/abs/2307.01163.pdf} {Improving language
  plasticity via pretraining with active forgetting}.
\newblock \emph{Advances in Neural Information Processing Systems},
  36:31543--31557.

\bibitem[{Conneau et~al.(2020{\natexlab{a}})Conneau, Khandelwal, Goyal,
  Chaudhary, Wenzek, Guzm{\'a}n, Grave, Ott, Zettlemoyer, and
  Stoyanov}]{conneau-etal-2020-unsupervised}
Alexis Conneau, Kartikay Khandelwal, Naman Goyal, Vishrav Chaudhary, Guillaume
  Wenzek, Francisco Guzm{\'a}n, Edouard Grave, Myle Ott, Luke Zettlemoyer, and
  Veselin Stoyanov. 2020{\natexlab{a}}.
\newblock \href {https://aclanthology.org/2020.acl-main.747} {Unsupervised
  cross-lingual representation learning at scale}.
\newblock In \emph{Proceedings of the 58th Annual Meeting of the Association
  for Computational Linguistics}, pages 8440--8451. Association for
  Computational Linguistics.

\bibitem[{Conneau et~al.(2020{\natexlab{b}})Conneau, Wu, Li, Zettlemoyer, and
  Stoyanov}]{conneau2020emerging}
Alexis Conneau, Shijie Wu, Haoran Li, Luke Zettlemoyer, and Veselin Stoyanov.
  2020{\natexlab{b}}.
\newblock \href {https://aclanthology.org/2020.acl-main.536} {Emerging
  cross-lingual structure in pretrained language models}.
\newblock In \emph{Proceedings of the 58th Annual Meeting of the Association
  for Computational Linguistics}, pages 6022--6034. Association for
  Computational Linguistics.

\bibitem[{Dufter and Sch{\"u}tze(2020)}]{dufter2020identifying}
Philipp Dufter and Hinrich Sch{\"u}tze. 2020.
\newblock \href {https://aclanthology.org/2020.emnlp-main.358} {Identifying
  elements essential for {BERT}{'}s multilinguality}.
\newblock In \emph{Proceedings of the 2020 Conference on Empirical Methods in
  Natural Language Processing (EMNLP)}, pages 4423--4437. Association for
  Computational Linguistics.

\bibitem[{Elhage et~al.(2021)Elhage, Nanda, Olsson, Henighan, Joseph, Mann,
  Askell, Bai, Chen, Conerly, DasSarma, Drain, Ganguli, Hatfield-Dodds,
  Hernandez, Jones, Kernion, Lovitt, Ndousse, Amodei, Brown, Clark, Kaplan,
  McCandlish, and Olah}]{elhage2021mathematical}
Nelson Elhage, Neel Nanda, Catherine Olsson, Tom Henighan, Nicholas Joseph, Ben
  Mann, Amanda Askell, Yuntao Bai, Anna Chen, Tom Conerly, Nova DasSarma, Dawn
  Drain, Deep Ganguli, Zac Hatfield-Dodds, Danny Hernandez, Andy Jones, Jackson
  Kernion, Liane Lovitt, Kamal Ndousse, Dario Amodei, Tom Brown, Jack Clark,
  Jared Kaplan, Sam McCandlish, and Chris Olah. 2021.
\newblock \href {https://transformer-circuits.pub/2021/framework/index.html} {A
  mathematical framework for transformer circuits}.
\newblock \emph{Transformer Circuits Thread}.

\bibitem[{Faysse et~al.(2024)Faysse, Fernandes, Guerreiro, Loison, Alves,
  Corro, Boizard, Alves, Rei, Martins et~al.}]{faysse2024croissantllm}
Manuel Faysse, Patrick Fernandes, Nuno Guerreiro, Ant{\'o}nio Loison, Duarte
  Alves, Caio Corro, Nicolas Boizard, Jo{\~a}o Alves, Ricardo Rei, Pedro
  Martins, et~al. 2024.
\newblock \href {https://arxiv.org/abs/2402.00786} {{CroissantLLM}: A truly
  bilingual french-english language model}.
\newblock \emph{arXiv preprint arXiv:2402.00786}.

\bibitem[{Feng et~al.(2022)Feng, Li, and Koehn}]{feng2022toward}
Yukun Feng, Feng Li, and Philipp Koehn. 2022.
\newblock \href {https://aclanthology.org/2022.emnlp-main.400} {Toward the
  limitation of code-switching in cross-lingual transfer}.
\newblock In \emph{Proceedings of the 2022 Conference on Empirical Methods in
  Natural Language Processing}, pages 5966--5971. Association for Computational
  Linguistics.

\bibitem[{French(1999)}]{french1999catastrophic}
Robert~M. French. 1999.
\newblock \href
  {https://www.sciencedirect.com/science/article/pii/S1364661399012942}
  {Catastrophic forgetting in connectionist networks}.
\newblock \emph{Trends in Cognitive Sciences}, 3(4):128--135.

\bibitem[{Gage(1994)}]{gage1994new}
Philip Gage. 1994.
\newblock A new algorithm for data compression.
\newblock \emph{C Users Journal}, 12(2):23--38.

\bibitem[{Gao et~al.(2020)Gao, Biderman, Black, Golding, Hoppe, Foster, Phang,
  He, Thite, Nabeshima, Presser, and Leahy}]{gao2020pile}
Leo Gao, Stella Biderman, Sid Black, Laurence Golding, Travis Hoppe, Charles
  Foster, Jason Phang, Horace He, Anish Thite, Noa Nabeshima, Shawn Presser,
  and Connor Leahy. 2020.
\newblock \href {https://arxiv.org/pdf/2101.00027.pdf} {The {Pile}: An 800{GB}
  dataset of diverse text for language modeling}.
\newblock \emph{arXiv preprint arXiv:2101.00027}.

\bibitem[{Hagberg et~al.(2008)Hagberg, Swart, and Chult}]{hagberg2008exploring}
Aric Hagberg, Pieter Swart, and Daniel Chult. 2008.
\newblock \href
  {https://www.researchgate.net/publication/236407765_Exploring_Network_Structure_Dynamics_and_Function_Using_NetworkX}
  {Exploring network structure, dynamics, and function using {NetworkX}}.
\newblock \emph{Proceedings of the 7th Python in Science Conference}.

\bibitem[{Hanna et~al.(2024)Hanna, Liu, and Variengien}]{hanna2024does}
Michael Hanna, Ollie Liu, and Alexandre Variengien. 2024.
\newblock \href {https://arxiv.org/abs/2305.00586.pdf} {How does {GPT}-2
  compute greater-than?: Interpreting mathematical abilities in a pre-trained
  language model}.
\newblock \emph{Advances in Neural Information Processing Systems}, 36.

\bibitem[{Hoffmann et~al.(2022)Hoffmann, Borgeaud, Mensch, Buchatskaya, Cai,
  Rutherford, de~las Casas, Hendricks, Welbl, Clark, Hennigan, Noland,
  Millican, van~den Driessche, Damoc, Guy, Osindero, Simonyan, Elsen, Vinyals,
  Rae, and Sifre}]{hoffmann2022chinchilla}
Jordan Hoffmann, Sebastian Borgeaud, Arthur Mensch, Elena Buchatskaya, Trevor
  Cai, Eliza Rutherford, Diego de~las Casas, Lisa~Anne Hendricks, Johannes
  Welbl, Aidan Clark, Tom Hennigan, Eric Noland, Katherine Millican, George
  van~den Driessche, Bogdan Damoc, Aurelia Guy, Simon Osindero, Karen Simonyan,
  Erich Elsen, Oriol Vinyals, Jack~William Rae, and Laurent Sifre. 2022.
\newblock \href {https://openreview.net/forum?id=iBBcRUlOAPR} {An empirical
  analysis of compute-optimal large language model training}.
\newblock In \emph{Advances in Neural Information Processing Systems}.

\bibitem[{Huang et~al.(2023)Huang, Zelikman, Chen, Wu, Valiant, and
  Liang}]{huang2023lexinvariant}
Qian Huang, Eric Zelikman, Sarah~Li Chen, Yuhuai Wu, Gregory Valiant, and Percy
  Liang. 2023.
\newblock \href {https://openreview.net/forum?id=NiQTy0NW1L} {Lexinvariant
  language models}.
\newblock In \emph{Thirty-seventh Conference on Neural Information Processing
  Systems}.

\bibitem[{K et~al.(2020)K, Wang, Mayhew, and Roth}]{wang2019cross}
Karthikeyan K, Zihan Wang, Stephen Mayhew, and Dan Roth. 2020.
\newblock \href {https://openreview.net/forum?id=HJeT3yrtDr} {Cross-lingual
  ability of multilingual {BERT}: An empirical study}.
\newblock In \emph{International Conference on Learning Representations}.

\bibitem[{Karp(1978)}]{Karp:M78/67}
Richard~M. Karp. 1978.
\newblock \href {http://www2.eecs.berkeley.edu/Pubs/TechRpts/1978/29160.html}
  {An algorithm to solve the mxn assignment problem in expected time {O}(mn log
  n)}.
\newblock Technical Report UCB/ERL M78/67, EECS Department, University of
  California, Berkeley.

\bibitem[{Kharitonov et~al.(2021)Kharitonov, Baroni, and
  Hupkes}]{kharitonov2021bpe}
Eugene Kharitonov, Marco Baroni, and Dieuwke Hupkes. 2021.
\newblock \href {https://arxiv.org/abs/2110.02782} {How {BPE} affects
  memorization in transformers}.
\newblock \emph{arXiv preprint arXiv:2110.02782}.

\bibitem[{Kingma and Ba(2015)}]{kingma2015adam}
Diederik Kingma and Jimmy Ba. 2015.
\newblock Adam: A method for stochastic optimization.
\newblock In \emph{International Conference on Learning Representations}, San
  Diego, CA, USA.

\bibitem[{Koehn(2005)}]{koehn2005europarl}
Philipp Koehn. 2005.
\newblock \href {https://aclanthology.org/2005.mtsummit-papers.11} {{E}uroparl:
  A parallel corpus for statistical machine translation}.
\newblock In \emph{Proceedings of Machine Translation Summit X: Papers}, pages
  79--86, Phuket, Thailand.

\bibitem[{K{\"o}pf et~al.(2023)K{\"o}pf, Kilcher, von R{\"u}tte, Anagnostidis,
  Tam, Stevens, Barhoum, Nguyen, Stanley, Nagyfi, ES, Suri, Glushkov,
  Dantuluri, Maguire, Schuhmann, Nguyen, and Mattick}]{kopf2023openassistant}
Andreas K{\"o}pf, Yannic Kilcher, Dimitri von R{\"u}tte, Sotiris Anagnostidis,
  Zhi~Rui Tam, Keith Stevens, Abdullah Barhoum, Duc~Minh Nguyen, Oliver
  Stanley, Rich{\'a}rd Nagyfi, Shahul ES, Sameer Suri, David~Alexandrovich
  Glushkov, Arnav~Varma Dantuluri, Andrew Maguire, Christoph Schuhmann, Huu
  Nguyen, and Alexander~Julian Mattick. 2023.
\newblock \href {https://openreview.net/forum?id=VSJotgbPHF} {{OpenAssistant}
  conversations - democratizing large language model alignment}.
\newblock In \emph{Thirty-seventh Conference on Neural Information Processing
  Systems Datasets and Benchmarks Track}.

\bibitem[{Kudo and Richardson(2018)}]{kudo2018sentencepiece}
Taku Kudo and John Richardson. 2018.
\newblock \href {https://arxiv.org/pdf/1808.06226.pdf} {{SentencePiece}: A
  simple and language independent subword tokenizer and detokenizer for neural
  text processing}.
\newblock In \emph{Proceedings of the 2018 Conference on Empirical Methods in
  Natural Language Processing: System Demonstrations}, pages 66--71.

\bibitem[{Lample and Conneau(2019)}]{lample2019cross}
Guillaume Lample and Alexis Conneau. 2019.
\newblock \href {https://arxiv.org/abs/1901.07291} {Cross-lingual language
  model pretraining}.
\newblock \emph{arXiv preprint arXiv:1901.07291}.

\bibitem[{McCloskey and Cohen(1989)}]{mccloskey1989catastrophic}
Michael McCloskey and Neal~J. Cohen. 1989.
\newblock \href
  {https://www.sciencedirect.com/science/article/pii/S0079742108605368}
  {Catastrophic interference in connectionist networks: The sequential learning
  problem}.
\newblock In \emph{Psychology of Learning and Motivation}, volume~24, pages
  109--165. Academic Press.

\bibitem[{Mikolov et~al.(2013)Mikolov, Chen, Corrado, and
  Dean}]{mikolov2013efficient}
Tom{\'{a}}s Mikolov, Kai Chen, Greg Corrado, and Jeffrey Dean. 2013.
\newblock \href {http://arxiv.org/abs/1301.3781} {Efficient estimation of word
  representations in vector space}.
\newblock In \emph{1st International Conference on Learning Representations,
  Workshop Track Proceedings}, Scottsdale, Arizona, USA.

\bibitem[{Pfeiffer et~al.(2022)Pfeiffer, Goyal, Lin, Li, Cross, Riedel, and
  Artetxe}]{pfeiffer-etal-2022-lifting}
Jonas Pfeiffer, Naman Goyal, Xi~Lin, Xian Li, James Cross, Sebastian Riedel,
  and Mikel Artetxe. 2022.
\newblock \href {https://aclanthology.org/2022.naacl-main.255} {Lifting the
  curse of multilinguality by pre-training modular transformers}.
\newblock In \emph{Proceedings of the 2022 Conference of the North American
  Chapter of the Association for Computational Linguistics: Human Language
  Technologies}, pages 3479--3495, Seattle, United States. Association for
  Computational Linguistics.

\bibitem[{Pires et~al.(2019)Pires, Schlinger, and
  Garrette}]{pires2019multilingual}
Telmo Pires, Eva Schlinger, and Dan Garrette. 2019.
\newblock \href {https://aclanthology.org/P19-1493.pdf} {How multilingual is
  multilingual {BERT}?}
\newblock In \emph{Proceedings of the 57th Annual Meeting of the Association
  for Computational Linguistics}, pages 4996--5001.

\bibitem[{{PleIAs}(2024)}]{french_pd_books_dataset}
{PleIAs}. 2024.
\newblock {French-PD-Books} dataset.
\newblock \url{https://huggingface.co/datasets/PleIAs/French-PD-Books}.
\newblock Accessed in 01/2024, Hugging Face Datasets library.

\bibitem[{Radford et~al.(2019)Radford, Wu, Child, Luan, Amodei, and
  Sutskever}]{radford2019language}
Alec Radford, Jeffrey Wu, Rewon Child, David Luan, Dario Amodei, and Ilya
  Sutskever. 2019.
\newblock \href
  {https://cdn.openai.com/better-language-models/language_models_are_unsupervised_multitask_learners.pdf}
  {Language models are unsupervised multitask learners}.
\newblock \emph{OpenAI Blog}.

\bibitem[{Reid and Artetxe(2022)}]{reid2022paradise}
Machel Reid and Mikel Artetxe. 2022.
\newblock \href {https://aclanthology.org/2022.naacl-main.58} {{PARADISE}:
  Exploiting parallel data for multilingual sequence-to-sequence pretraining}.
\newblock In \emph{Proceedings of the 2022 Conference of the North American
  Chapter of the Association for Computational Linguistics: Human Language
  Technologies}, pages 800--810, Seattle, United States. Association for
  Computational Linguistics.

\bibitem[{Scao et~al.(2023)Scao, Fan, Akiki, Pavlick, Ilić, Hesslow,
  Castagné, Luccioni, Yvon, Gallé, Tow, Rush, Biderman, Webson, Ammanamanchi,
  Wang, Sagot, Muennighoff, del Moral, Ruwase, Bawden, Bekman, McMillan-Major,
  Beltagy, Nguyen, Saulnier, Tan, Suarez, Sanh, Laurençon, Jernite, Launay,
  Mitchell, Raffel, Gokaslan, Simhi, Soroa, Aji, Alfassy, Rogers, Nitzav, Xu,
  Mou, Emezue, Klamm, Leong, van Strien, Adelani, Radev, Ponferrada, Levkovizh,
  Kim, Natan, Toni, Dupont, Kruszewski, Pistilli, Elsahar, Benyamina, Tran, Yu,
  Abdulmumin, Johnson, Gonzalez-Dios, de~la Rosa, Chim, Dodge, Zhu, Chang,
  Frohberg, Tobing, Bhattacharjee, Almubarak, Chen, Lo, Werra, Weber, Phan,
  allal, Tanguy, Dey, Muñoz, Masoud, Grandury, Šaško, Huang, Coavoux, Singh,
  Jiang, Vu, Jauhar, Ghaleb, Subramani, Kassner, Khamis, Nguyen, Espejel,
  de~Gibert, Villegas, Henderson, Colombo, Amuok, Lhoest, Harliman, Bommasani,
  López, Ribeiro, Osei, Pyysalo, Nagel, Bose, Muhammad, Sharma, Longpre,
  Nikpoor, Silberberg, Pai, Zink, Torrent, Schick, Thrush, Danchev, Nikoulina,
  Laippala, Lepercq, Prabhu, Alyafeai, Talat, Raja, Heinzerling, Si, Taşar,
  Salesky, Mielke, Lee, Sharma, Santilli, Chaffin, Stiegler, Datta, Szczechla,
  Chhablani, Wang, Pandey, Strobelt, Fries, Rozen, Gao, Sutawika, Bari,
  Al-shaibani, Manica, Nayak, Teehan, Albanie, Shen, Ben-David, Bach, Kim,
  Bers, Fevry, Neeraj, Thakker, Raunak, Tang, Yong, Sun, Brody, Uri, Tojarieh,
  Roberts, Chung, Tae, Phang, Press, Li, Narayanan, Bourfoune, Casper, Rasley,
  Ryabinin, Mishra, Zhang, Shoeybi, Peyrounette, Patry, Tazi, Sanseviero, von
  Platen, Cornette, Lavallée, Lacroix, Rajbhandari, Gandhi, Smith, Requena,
  Patil, Dettmers, Baruwa, Singh, Cheveleva, Ligozat, Subramonian, Névéol,
  Lovering, Garrette, Tunuguntla, Reiter, Taktasheva, Voloshina, Bogdanov,
  Winata, Schoelkopf, Kalo, Novikova, Forde, Clive, Kasai, Kawamura, Hazan,
  Carpuat, Clinciu, Kim, Cheng, Serikov, Antverg, van~der Wal, Zhang, Zhang,
  Gehrmann, Mirkin, Pais, Shavrina, Scialom, Yun, Limisiewicz, Rieser,
  Protasov, Mikhailov, Pruksachatkun, Belinkov, Bamberger, Kasner, Rueda,
  Pestana, Feizpour, Khan, Faranak, Santos, Hevia, Unldreaj, Aghagol,
  Abdollahi, Tammour, HajiHosseini, Behroozi, Ajibade, Saxena, Ferrandis,
  McDuff, Contractor, Lansky, David, Kiela, Nguyen, Tan, Baylor, Ozoani, Mirza,
  Ononiwu, Rezanejad, Jones, Bhattacharya, Solaiman, Sedenko, Nejadgholi,
  Passmore, Seltzer, Sanz, Dutra, Samagaio, Elbadri, Mieskes, Gerchick,
  Akinlolu, McKenna, Qiu, Ghauri, Burynok, Abrar, Rajani, Elkott, Fahmy,
  Samuel, An, Kromann, Hao, Alizadeh, Shubber, Wang, Roy, Viguier, Le, Oyebade,
  Le, Yang, Nguyen, Kashyap, Palasciano, Callahan, Shukla, Miranda-Escalada,
  Singh, Beilharz, Wang, Brito, Zhou, Jain, Xu, Fourrier, Periñán, Molano,
  Yu, Manjavacas, Barth, Fuhrimann, Altay, Bayrak, Burns, Vrabec, Bello, Dash,
  Kang, Giorgi, Golde, Posada, Sivaraman, Bulchandani, Liu, Shinzato,
  de~Bykhovetz, Takeuchi, Pàmies, Castillo, Nezhurina, Sänger, Samwald,
  Cullan, Weinberg, Wolf, Mihaljcic, Liu, Freidank, Kang, Seelam, Dahlberg,
  Broad, Muellner, Fung, Haller, Chandrasekhar, Eisenberg, Martin, Canalli, Su,
  Su, Cahyawijaya, Garda, Deshmukh, Mishra, Kiblawi, Ott, Sang-aroonsiri,
  Kumar, Schweter, Bharati, Laud, Gigant, Kainuma, Kusa, Labrak, Bajaj,
  Venkatraman, Xu, Xu, Xu, Tan, Xie, Ye, Bras, Belkada, and
  Wolf}]{workshop2023bloom}
Teven~Le Scao, Angela Fan, Christopher Akiki, Ellie Pavlick, Suzana Ilić,
  Daniel Hesslow, Roman Castagné, Alexandra~Sasha Luccioni, François Yvon,
  Matthias Gallé, Jonathan Tow, Alexander~M. Rush, Stella Biderman, Albert
  Webson, Pawan~Sasanka Ammanamanchi, Thomas Wang, Benoît Sagot, Niklas
  Muennighoff, Albert~Villanova del Moral, Olatunji Ruwase, Rachel Bawden, Stas
  Bekman, Angelina McMillan-Major, Iz~Beltagy, Huu Nguyen, Lucile Saulnier,
  Samson Tan, Pedro~Ortiz Suarez, Victor Sanh, Hugo Laurençon, Yacine Jernite,
  Julien Launay, Margaret Mitchell, Colin Raffel, Aaron Gokaslan, Adi Simhi,
  Aitor Soroa, Alham~Fikri Aji, Amit Alfassy, Anna Rogers, Ariel~Kreisberg
  Nitzav, Canwen Xu, Chenghao Mou, Chris Emezue, Christopher Klamm, Colin
  Leong, Daniel van Strien, David~Ifeoluwa Adelani, Dragomir Radev,
  Eduardo~González Ponferrada, Efrat Levkovizh, Ethan Kim, Eyal~Bar Natan,
  Francesco~De Toni, Gérard Dupont, Germán Kruszewski, Giada Pistilli, Hady
  Elsahar, Hamza Benyamina, Hieu Tran, Ian Yu, Idris Abdulmumin, Isaac Johnson,
  Itziar Gonzalez-Dios, Javier de~la Rosa, Jenny Chim, Jesse Dodge, Jian Zhu,
  Jonathan Chang, Jörg Frohberg, Joseph Tobing, Joydeep Bhattacharjee, Khalid
  Almubarak, Kimbo Chen, Kyle Lo, Leandro~Von Werra, Leon Weber, Long Phan,
  Loubna~Ben allal, Ludovic Tanguy, Manan Dey, Manuel~Romero Muñoz, Maraim
  Masoud, María Grandury, Mario Šaško, Max Huang, Maximin Coavoux, Mayank
  Singh, Mike Tian-Jian Jiang, Minh~Chien Vu, Mohammad~A. Jauhar, Mustafa
  Ghaleb, Nishant Subramani, Nora Kassner, Nurulaqilla Khamis, Olivier Nguyen,
  Omar Espejel, Ona de~Gibert, Paulo Villegas, Peter Henderson, Pierre Colombo,
  Priscilla Amuok, Quentin Lhoest, Rheza Harliman, Rishi Bommasani,
  Roberto~Luis López, Rui Ribeiro, Salomey Osei, Sampo Pyysalo, Sebastian
  Nagel, Shamik Bose, Shamsuddeen~Hassan Muhammad, Shanya Sharma, Shayne
  Longpre, Somaieh Nikpoor, Stanislav Silberberg, Suhas Pai, Sydney Zink,
  Tiago~Timponi Torrent, Timo Schick, Tristan Thrush, Valentin Danchev,
  Vassilina Nikoulina, Veronika Laippala, Violette Lepercq, Vrinda Prabhu, Zaid
  Alyafeai, Zeerak Talat, Arun Raja, Benjamin Heinzerling, Chenglei Si,
  Davut~Emre Taşar, Elizabeth Salesky, Sabrina~J. Mielke, Wilson~Y. Lee,
  Abheesht Sharma, Andrea Santilli, Antoine Chaffin, Arnaud Stiegler, Debajyoti
  Datta, Eliza Szczechla, Gunjan Chhablani, Han Wang, Harshit Pandey, Hendrik
  Strobelt, Jason~Alan Fries, Jos Rozen, Leo Gao, Lintang Sutawika, M~Saiful
  Bari, Maged~S. Al-shaibani, Matteo Manica, Nihal Nayak, Ryan Teehan, Samuel
  Albanie, Sheng Shen, Srulik Ben-David, Stephen~H. Bach, Taewoon Kim, Tali
  Bers, Thibault Fevry, Trishala Neeraj, Urmish Thakker, Vikas Raunak, Xiangru
  Tang, Zheng-Xin Yong, Zhiqing Sun, Shaked Brody, Yallow Uri, Hadar Tojarieh,
  Adam Roberts, Hyung~Won Chung, Jaesung Tae, Jason Phang, Ofir Press, Conglong
  Li, Deepak Narayanan, Hatim Bourfoune, Jared Casper, Jeff Rasley, Max
  Ryabinin, Mayank Mishra, Minjia Zhang, Mohammad Shoeybi, Myriam Peyrounette,
  Nicolas Patry, Nouamane Tazi, Omar Sanseviero, Patrick von Platen, Pierre
  Cornette, Pierre~François Lavallée, Rémi Lacroix, Samyam Rajbhandari,
  Sanchit Gandhi, Shaden Smith, Stéphane Requena, Suraj Patil, Tim Dettmers,
  Ahmed Baruwa, Amanpreet Singh, Anastasia Cheveleva, Anne-Laure Ligozat, Arjun
  Subramonian, Aurélie Névéol, Charles Lovering, Dan Garrette, Deepak
  Tunuguntla, Ehud Reiter, Ekaterina Taktasheva, Ekaterina Voloshina, Eli
  Bogdanov, Genta~Indra Winata, Hailey Schoelkopf, Jan-Christoph Kalo,
  Jekaterina Novikova, Jessica~Zosa Forde, Jordan Clive, Jungo Kasai, Ken
  Kawamura, Liam Hazan, Marine Carpuat, Miruna Clinciu, Najoung Kim, Newton
  Cheng, Oleg Serikov, Omer Antverg, Oskar van~der Wal, Rui Zhang, Ruochen
  Zhang, Sebastian Gehrmann, Shachar Mirkin, Shani Pais, Tatiana Shavrina,
  Thomas Scialom, Tian Yun, Tomasz Limisiewicz, Verena Rieser, Vitaly Protasov,
  Vladislav Mikhailov, Yada Pruksachatkun, Yonatan Belinkov, Zachary Bamberger,
  Zdeněk Kasner, Alice Rueda, Amanda Pestana, Amir Feizpour, Ammar Khan, Amy
  Faranak, Ana Santos, Anthony Hevia, Antigona Unldreaj, Arash Aghagol, Arezoo
  Abdollahi, Aycha Tammour, Azadeh HajiHosseini, Bahareh Behroozi, Benjamin
  Ajibade, Bharat Saxena, Carlos~Muñoz Ferrandis, Daniel McDuff, Danish
  Contractor, David Lansky, Davis David, Douwe Kiela, Duong~A. Nguyen, Edward
  Tan, Emi Baylor, Ezinwanne Ozoani, Fatima Mirza, Frankline Ononiwu, Habib
  Rezanejad, Hessie Jones, Indrani Bhattacharya, Irene Solaiman, Irina Sedenko,
  Isar Nejadgholi, Jesse Passmore, Josh Seltzer, Julio~Bonis Sanz, Livia Dutra,
  Mairon Samagaio, Maraim Elbadri, Margot Mieskes, Marissa Gerchick, Martha
  Akinlolu, Michael McKenna, Mike Qiu, Muhammed Ghauri, Mykola Burynok, Nafis
  Abrar, Nazneen Rajani, Nour Elkott, Nour Fahmy, Olanrewaju Samuel, Ran An,
  Rasmus Kromann, Ryan Hao, Samira Alizadeh, Sarmad Shubber, Silas Wang, Sourav
  Roy, Sylvain Viguier, Thanh Le, Tobi Oyebade, Trieu Le, Yoyo Yang, Zach
  Nguyen, Abhinav~Ramesh Kashyap, Alfredo Palasciano, Alison Callahan, Anima
  Shukla, Antonio Miranda-Escalada, Ayush Singh, Benjamin Beilharz, Bo~Wang,
  Caio Brito, Chenxi Zhou, Chirag Jain, Chuxin Xu, Clémentine Fourrier,
  Daniel~León Periñán, Daniel Molano, Dian Yu, Enrique Manjavacas, Fabio
  Barth, Florian Fuhrimann, Gabriel Altay, Giyaseddin Bayrak, Gully Burns,
  Helena~U. Vrabec, Imane Bello, Ishani Dash, Jihyun Kang, John Giorgi, Jonas
  Golde, Jose~David Posada, Karthik~Rangasai Sivaraman, Lokesh Bulchandani,
  Lu~Liu, Luisa Shinzato, Madeleine~Hahn de~Bykhovetz, Maiko Takeuchi, Marc
  Pàmies, Maria~A Castillo, Marianna Nezhurina, Mario Sänger, Matthias
  Samwald, Michael Cullan, Michael Weinberg, Michiel~De Wolf, Mina Mihaljcic,
  Minna Liu, Moritz Freidank, Myungsun Kang, Natasha Seelam, Nathan Dahlberg,
  Nicholas~Michio Broad, Nikolaus Muellner, Pascale Fung, Patrick Haller, Ramya
  Chandrasekhar, Renata Eisenberg, Robert Martin, Rodrigo Canalli, Rosaline Su,
  Ruisi Su, Samuel Cahyawijaya, Samuele Garda, Shlok~S Deshmukh, Shubhanshu
  Mishra, Sid Kiblawi, Simon Ott, Sinee Sang-aroonsiri, Srishti Kumar, Stefan
  Schweter, Sushil Bharati, Tanmay Laud, Théo Gigant, Tomoya Kainuma, Wojciech
  Kusa, Yanis Labrak, Yash~Shailesh Bajaj, Yash Venkatraman, Yifan Xu, Yingxin
  Xu, Yu~Xu, Zhe Tan, Zhongli Xie, Zifan Ye, Mathilde Bras, Younes Belkada, and
  Thomas Wolf. 2023.
\newblock \href {http://arxiv.org/abs/2211.05100} {{BLOOM}: A 176{B}-parameter
  open-access multilingual language model}.
\newblock \emph{arXiv preprint arXiv:2211.05100}.

\bibitem[{Schäfer et~al.(2024)Schäfer, Hofmann, Schlag, and
  Pimentel}]{schafer2024on}
Anton Schäfer, Thomas Hofmann, Imanol Schlag, and Tiago Pimentel. 2024.
\newblock \href {http://arxiv.org/abs/2404.06508} {On the effect of (near)
  duplicate subwords in language modelling}.
\newblock \emph{arXiv preprint arXiv:2404.06508}.

\bibitem[{Sennrich et~al.(2016)Sennrich, Haddow, and
  Birch}]{sennrich-etal-2016-neural}
Rico Sennrich, Barry Haddow, and Alexandra Birch. 2016.
\newblock \href {https://aclanthology.org/P16-1162} {Neural machine translation
  of rare words with subword units}.
\newblock In \emph{Proceedings of the 54th Annual Meeting of the Association
  for Computational Linguistics (Volume 1: Long Papers)}, pages 1715--1725,
  Berlin, Germany. Association for Computational Linguistics.

\bibitem[{Stani{\'c} et~al.(2023)Stani{\'c}, Ashley, Serikov, Kirsch, Faccio,
  Schmidhuber, Hofmann, and Schlag}]{stanic2023languini}
Aleksandar Stani{\'c}, Dylan Ashley, Oleg Serikov, Louis Kirsch, Francesco
  Faccio, J{\"u}rgen Schmidhuber, Thomas Hofmann, and Imanol Schlag. 2023.
\newblock \href {https://arxiv.org/abs/2309.11197} {The languini kitchen:
  Enabling language modelling research at different scales of compute}.
\newblock \emph{arXiv preprint arXiv:2309.11197}.

\bibitem[{Touvron et~al.(2023{\natexlab{a}})Touvron, Lavril, Izacard, Martinet,
  Lachaux, Lacroix, Rozière, Goyal, Hambro, Azhar, Rodriguez, Joulin, Grave,
  and Lample}]{touvron2023llama}
Hugo Touvron, Thibaut Lavril, Gautier Izacard, Xavier Martinet, Marie-Anne
  Lachaux, Timothée Lacroix, Baptiste Rozière, Naman Goyal, Eric Hambro,
  Faisal Azhar, Aurelien Rodriguez, Armand Joulin, Edouard Grave, and Guillaume
  Lample. 2023{\natexlab{a}}.
\newblock \href {https://arxiv.org/abs/2302.13971} {{LLaMA}: Open and efficient
  foundation language models}.
\newblock \emph{arXiv preprint arXiv:2302.13971}.

\bibitem[{Touvron et~al.(2023{\natexlab{b}})Touvron, Martin, Stone, Albert,
  Almahairi, Babaei, Bashlykov, Batra, Bhargava, Bhosale, Bikel, Blecher,
  Ferrer, Chen, Cucurull, Esiobu, Fernandes, Fu, Fu, Fuller, Gao, Goswami,
  Goyal, Hartshorn, Hosseini, Hou, Inan, Kardas, Kerkez, Khabsa, Kloumann,
  Korenev, Koura, Lachaux, Lavril, Lee, Liskovich, Lu, Mao, Martinet, Mihaylov,
  Mishra, Molybog, Nie, Poulton, Reizenstein, Rungta, Saladi, Schelten, Silva,
  Smith, Subramanian, Tan, Tang, Taylor, Williams, Kuan, Xu, Yan, Zarov, Zhang,
  Fan, Kambadur, Narang, Rodriguez, Stojnic, Edunov, and
  Scialom}]{touvron2023llama2}
Hugo Touvron, Louis Martin, Kevin Stone, Peter Albert, Amjad Almahairi, Yasmine
  Babaei, Nikolay Bashlykov, Soumya Batra, Prajjwal Bhargava, Shruti Bhosale,
  Dan Bikel, Lukas Blecher, Cristian~Canton Ferrer, Moya Chen, Guillem
  Cucurull, David Esiobu, Jude Fernandes, Jeremy Fu, Wenyin Fu, Brian Fuller,
  Cynthia Gao, Vedanuj Goswami, Naman Goyal, Anthony Hartshorn, Saghar
  Hosseini, Rui Hou, Hakan Inan, Marcin Kardas, Viktor Kerkez, Madian Khabsa,
  Isabel Kloumann, Artem Korenev, Punit~Singh Koura, Marie-Anne Lachaux,
  Thibaut Lavril, Jenya Lee, Diana Liskovich, Yinghai Lu, Yuning Mao, Xavier
  Martinet, Todor Mihaylov, Pushkar Mishra, Igor Molybog, Yixin Nie, Andrew
  Poulton, Jeremy Reizenstein, Rashi Rungta, Kalyan Saladi, Alan Schelten, Ruan
  Silva, Eric~Michael Smith, Ranjan Subramanian, Xiaoqing~Ellen Tan, Binh Tang,
  Ross Taylor, Adina Williams, Jian~Xiang Kuan, Puxin Xu, Zheng Yan, Iliyan
  Zarov, Yuchen Zhang, Angela Fan, Melanie Kambadur, Sharan Narang, Aurelien
  Rodriguez, Robert Stojnic, Sergey Edunov, and Thomas Scialom.
  2023{\natexlab{b}}.
\newblock \href {https://arxiv.org/pdf/2307.09288.pdf} {Llama 2: Open
  foundation and fine-tuned chat models}.
\newblock \emph{arXiv preprint arXiv:2307.09288}.

\bibitem[{Wang et~al.(2019)Wang, Singh, Michael, Hill, Levy, and
  Bowman}]{wang2019glue}
Alex Wang, Amanpreet Singh, Julian Michael, Felix Hill, Omer Levy, and
  Samuel~R. Bowman. 2019.
\newblock \href {https://openreview.net/forum?id=rJ4km2R5t7} {{GLUE}: A
  multi-task benchmark and analysis platform for natural language
  understanding}.
\newblock In \emph{International Conference on Learning Representations}.

\bibitem[{Wang et~al.(2020)Wang, Lipton, and Tsvetkov}]{wang2020negative}
Zirui Wang, Zachary~C Lipton, and Yulia Tsvetkov. 2020.
\newblock \href {https://aclanthology.org/2020.emnlp-main.359.pdf} {On negative
  interference in multilingual models: Findings and a meta-learning treatment}.
\newblock In \emph{Proceedings of the 2020 Conference on Empirical Methods in
  Natural Language Processing (EMNLP)}, pages 4438--4450.

\bibitem[{Wendler et~al.(2024)Wendler, Veselovsky, Monea, and
  West}]{wendler2024llamas}
Chris Wendler, Veniamin Veselovsky, Giovanni Monea, and Robert West. 2024.
\newblock \href {http://arxiv.org/abs/2402.10588} {Do {Llamas} work in
  {E}nglish? {O}n the latent language of multilingual transformers}.
\newblock \emph{arXiv preprint arXiv:2402.10588}.

\bibitem[{Wu and Dredze(2019)}]{wu2019beto}
Shijie Wu and Mark Dredze. 2019.
\newblock \href {https://aclanthology.org/D19-1077.pdf} {Beto, bentz, becas:
  The surprising cross-lingual effectiveness of {BERT}}.
\newblock In \emph{2019 Conference on Empirical Methods in Natural Language
  Processing and 9th International Joint Conference on Natural Language
  Processing, EMNLP-IJCNLP 2019}, pages 833--844.

\bibitem[{Ye et~al.(2023)Ye, Tao, and Kong}]{ye2023language}
Jiacheng Ye, Xijia Tao, and Lingpeng Kong. 2023.
\newblock \href {https://arxiv.org/abs/2306.06688.pdf} {Language versatilists
  vs. specialists: An empirical revisiting on multilingual transfer ability}.
\newblock \emph{arXiv preprint arXiv:2306.06688}.

\end{thebibliography}

\clearpage
\onecolumn
\appendix

\section{Experimental Setup}

\label{app:setup}

\paragraph{Model.}
We use a GPT-2-style decoder-only transformer architecture in our experiments \citep{radford2019language}.
Unless otherwise noted, we instantiate our model with 12 layers and a hidden size of 768, which results in 85M non-embedding parameters; this corresponds to Languini's \texttt{gpt-small} configuration.
We follow previous work and train our models with sequence length 512, batch size 128, the Adam optimiser \citep{kingma2015adam}, and a cosine learning rate schedule from 6e-4 to 6e-6 with 500 warmup steps.

\paragraph{Data.}
For the English settings, we use Languini's default datasets to train and evaluate our models. 
These are English books from a filtered version of the books3 subset from the Pile \citep{gao2020pile}.
The train set consists of a total of 23.9B tokens, while the test set contains i.i.d. books, with a total of roughly 11M tokens.
This data is tokenised into a vocabulary of size 16k, obtained using a BPE tokeniser trained with SentencePiece \citep{gage1994new,sennrich-etal-2016-neural,kudo2018sentencepiece}.
For our experiments in French, we use the French-PD-Books dataset \citep{french_pd_books_dataset}, to which we apply the preprocessing pipeline of the Languini Kitchen, but for French.
We train a separate BPE tokeniser on this French dataset, using a 16k-sized vocabulary.
Depending on the experiment, the French and English vocabularies are either kept separate (disjoint) or merged (anchored).
Unless otherwise noted, we train our models for 18,265 steps---i.e., the first 1.2B tokens in our dataset; this corresponds to a GPT small model trained for 6h on an RTX 3090 GPU, the Languini GPT small 6h setting \citep{stanic2023languini}. For experiments where we compare hidden representations or gradients on parallel French--English or cloned English sequences, we use data from the Europarl parallel corpus \citep{koehn2005europarl}.\looseness=-1

\paragraph{Evaluation.}

When evaluating $\ppl$ (from which we also compute $\mlpe$, $\monotok$ and $\deff$) on the held-out test set, we want to
ensure sufficient context for all predictions. To this end, we use a sliding window with steps of 128: we fill in a 512 tokens context, ignore the model's outputs on the initial 384, and evaluate it only using the last 128 tokens.

\clearpage
\section{Fitted Scaling Laws}

\label{app:appl_scaling_laws}
\begin{figure}[h]
    \centering
        \includegraphics[width=\linewidth]{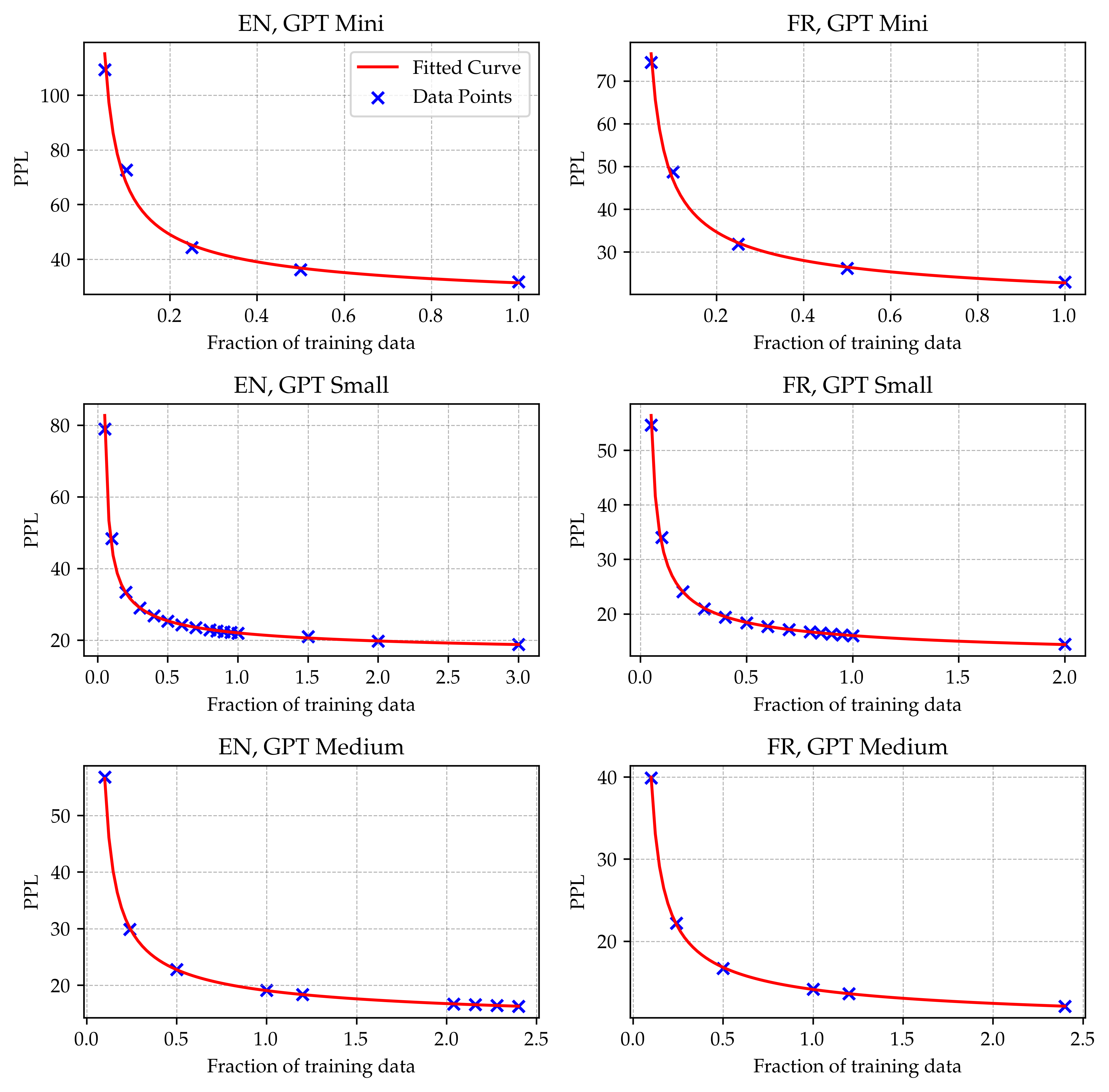}
    \caption{Fitted power laws curves predicting perplexity depending on the fraction of training tokens (compared to our standard 1.2B tokens) for different languages and model sizes.}
    \label{fig:scaling_laws_fit}
\end{figure}

To predict the performance of monolingual models depending on the amount of tokens they are trained on, we fit a power law curve to predict the relationship between number of training tokens and perplexity for models of all three sizes and for both languages (see \cref{fig:scaling_laws_fit}).

\clearpage
\section{Alignment of $\english_1$ and $\english_2$ Representations}
\label{app:alignment_en12}
\begin{figure}[h]
    \centering
        \includegraphics[width=0.85\linewidth]{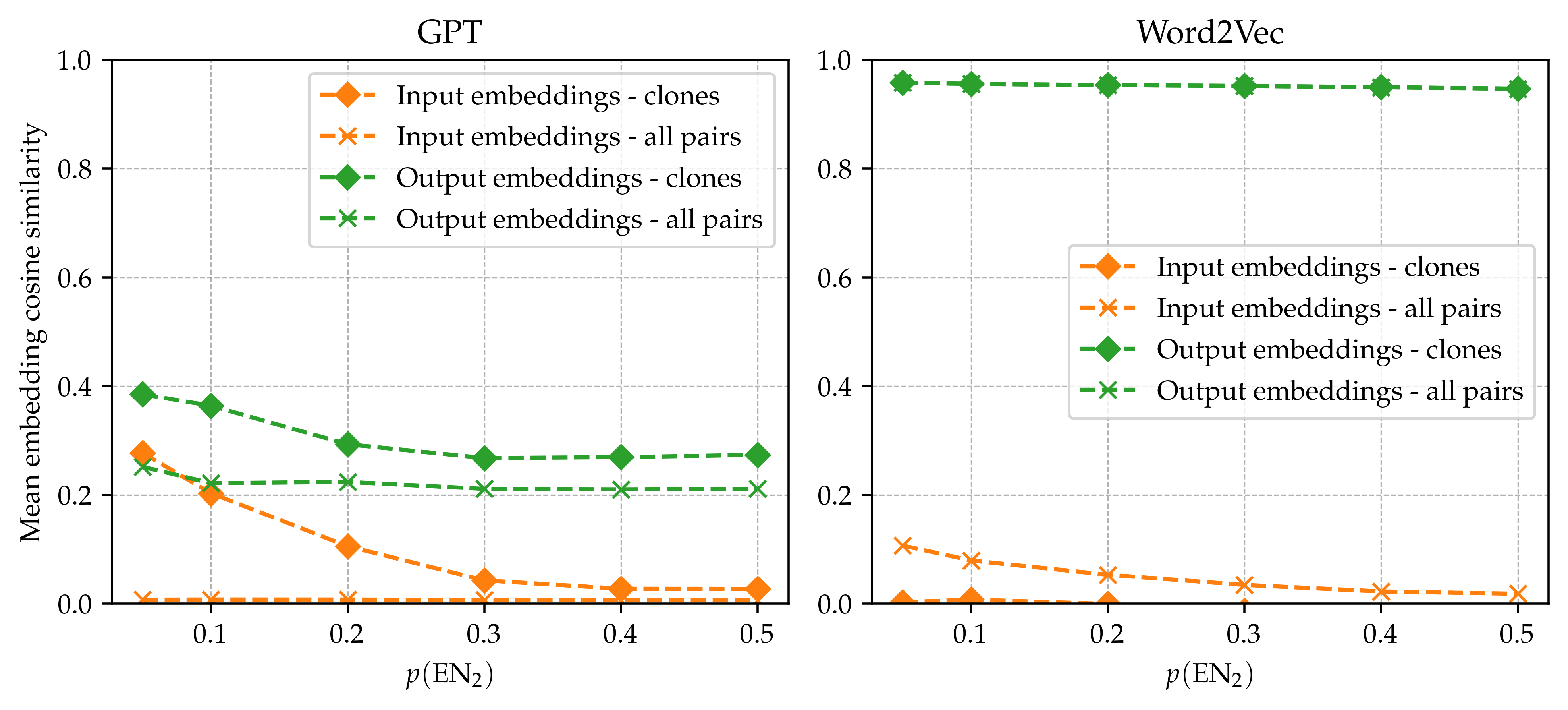}
    \caption{Embedding cosine similarity of corresponding duplicate subwords from $\english_1$ and $\english_2$ and random pairs to control for anisotropy. 
    Left: our GPT model. Right: Word2vec embeddings trained on the same data (computed with Gensim).}
    \label{fig:cossim_synth_ratio}
\end{figure}

\begin{figure}[h]
    \centering
        \includegraphics[width=0.55\linewidth]{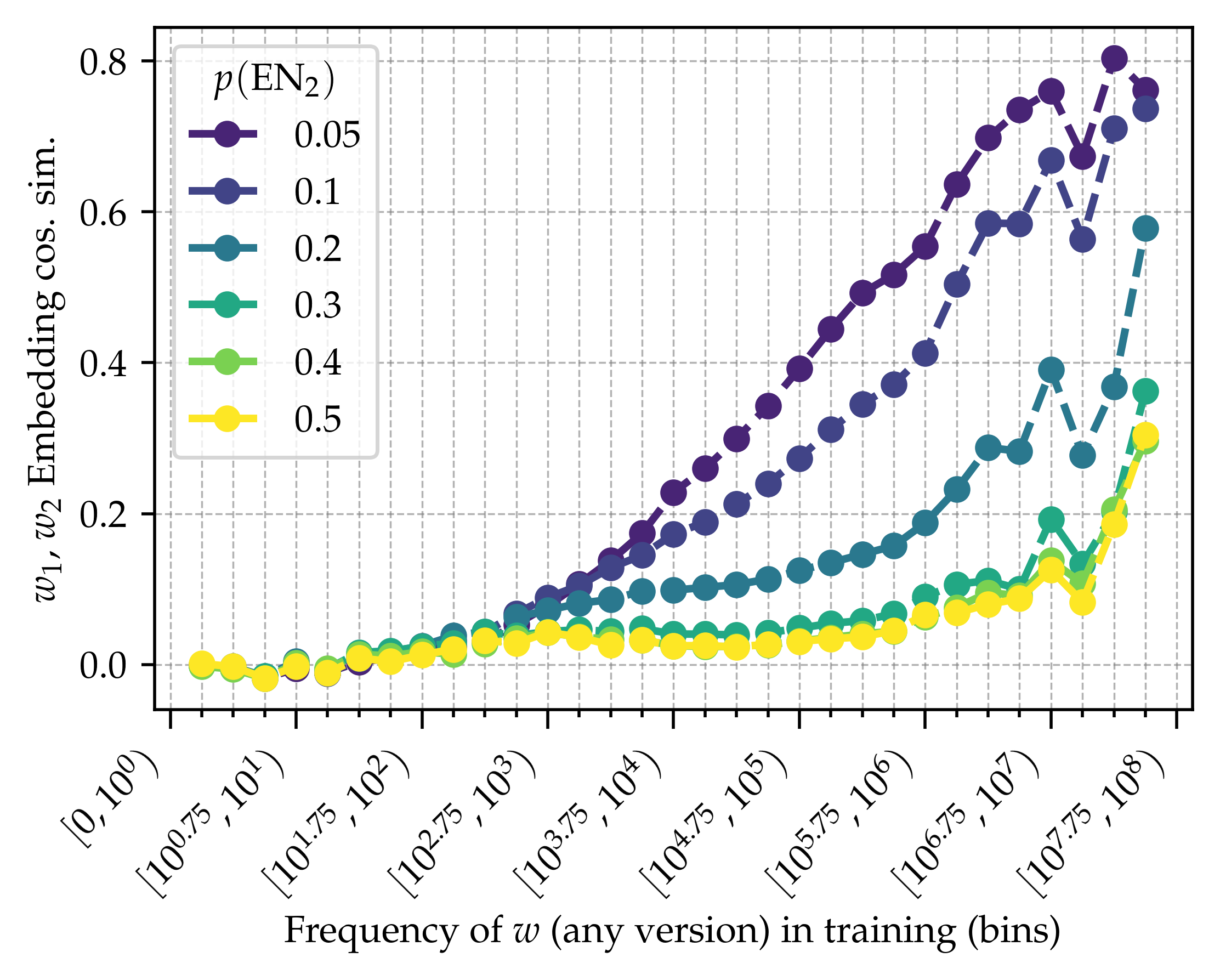}
    \caption{Embedding cosine similarity of corresponding cloned subwords $w_1 \cloneequiv w_2$ from $\english_1$ and $\english_2$, by frequency.
    }
    \label{fig:cossim_by_freq_synth_ratio}
\end{figure}

While, under balanced language sampling, embeddings of corresponding subwords are not much more aligned than embeddings of random pairs, we observe an increase in cosine similarity with increasing language imbalance: from 0.02 for \sfrac{50}{50} to 0.28 for \sfrac{95}{5} (see  \cref{fig:cossim_synth_ratio}). \cref{fig:cossim_by_freq_synth_ratio} shows that this alignment is higher for frequent subwords. This seems natural: at initialisation, subword embeddings are random and not aligned. Then, they become more and more aligned over the course of training.

Interestingly, the embeddings of a simple word2vec \citep{mikolov2013efficient} model do not show stronger alignment under higher imbalance. This might be due to a lack of shared parameters between the languages \citep{conneau2020emerging}.


\clearpage
\section{Anchor Points}
\label{app:full_anchors}

\subsection{Anchors on Cloned Languages}
\label{app:en12_anchors}

\begin{figure}[h]
    \centering
    \includegraphics[width=0.55\linewidth]{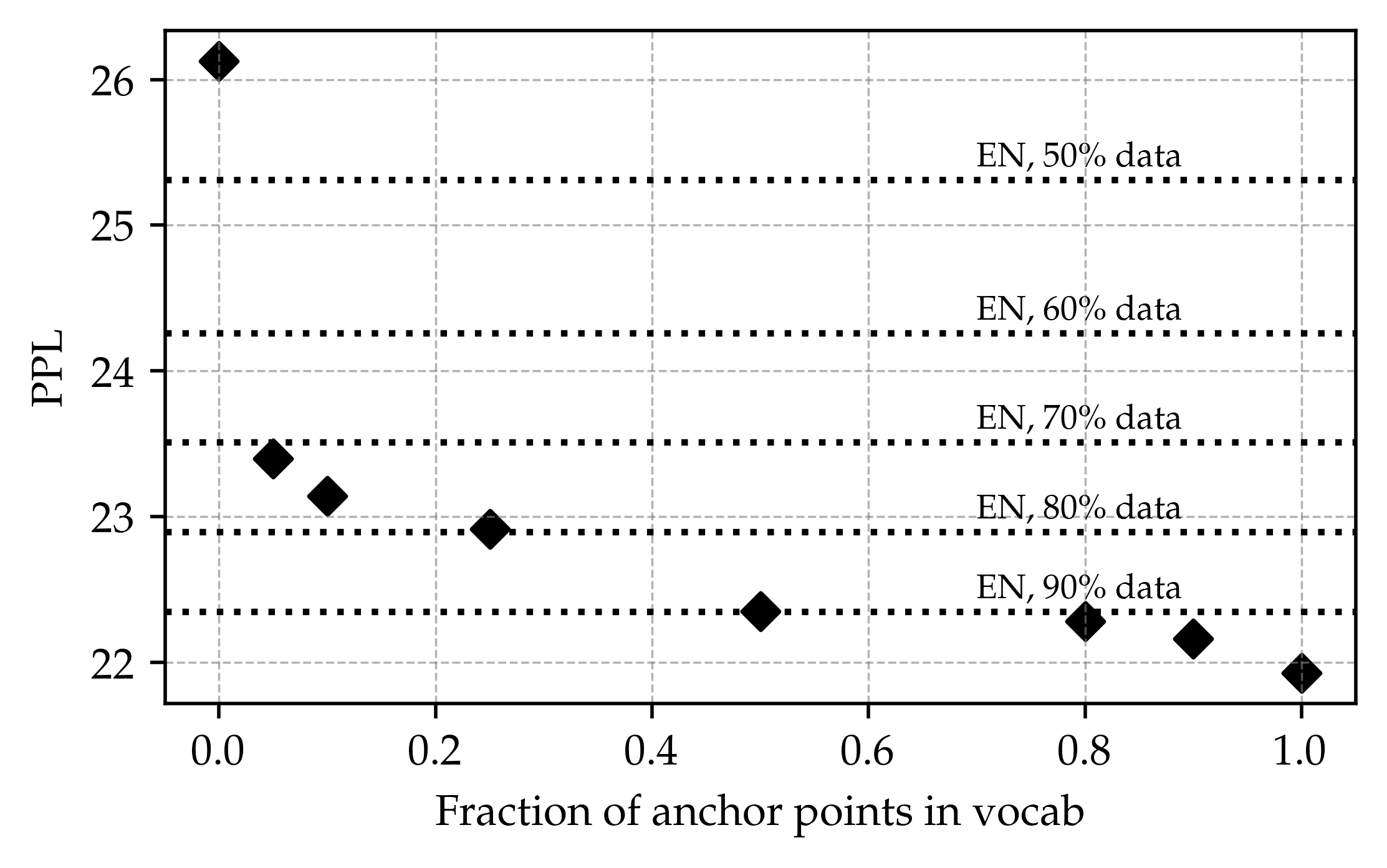}
    \caption{Perplexity by percentage of anchor points, i.e., overlap between $\english_1$ and $\english_2$ vocabularies. Models trained on balanced $\english_1/\english_2$ split.
    }
    \label{fig:ppl_anchor_points}
\end{figure}

As described earlier, previous works found that anchor points---i.e., lexical  items which overlap between languages---can lead to better generalisation and alignment of representations \citep{dufter2020identifying, pires2019multilingual, wu2019beto}.
In our cloned setting, we can investigate this in a controlled manner by varying the number of vocabulary elements we duplicate. While in the experiments described above we created $\english_2$ by duplicating the entire vocabulary, we now duplicate only a fraction. 
The remaining vocabulary is shared between $\english_1$ and $\english_2$.
In this experiment, we observe that a small number of anchor points already significantly boosts model performance (see \cref{fig:ppl_anchor_points}), which indicates improved generalisation.

\subsection{Anchors on Real Languages}
\label{sec:enfr_anchors}

English and French vocabularies naturally overlap, having common subwords. 
These shared elements potentially act as anchors, facilitating better cross-lingual generalisation.
However, the effectiveness of such anchor points may be moderated by semantic differences; for instance, a shared subword might carry a different meaning or connotations in English and French, affecting its utility as an anchor. 
Despite these nuances, anchor points appear to boost generalisation between real languages: when we merge the $\english$ and $\french$ vocabularies, we obtain better performance on both languages (compare \cref{tab:enfr_basic}, row 7 vs 11) as well as higher alignment of gradients (see \cref{app:gradient_similarity}). This aligns with our findings from the cloned language setting (see \cref{app:en12_anchors}). Given these benefits, it is natural to use an anchored (i.e., merged) vocabulary when possible.\footnote{In practice, this is usually achieved by training a tokeniser on multilingual data, instead of merging monolingually trained vocabularies.}

\clearpage

\section{Larger Models and More Data}
\begin{figure}[h]
    \centering
    \includegraphics[width=1.\linewidth]{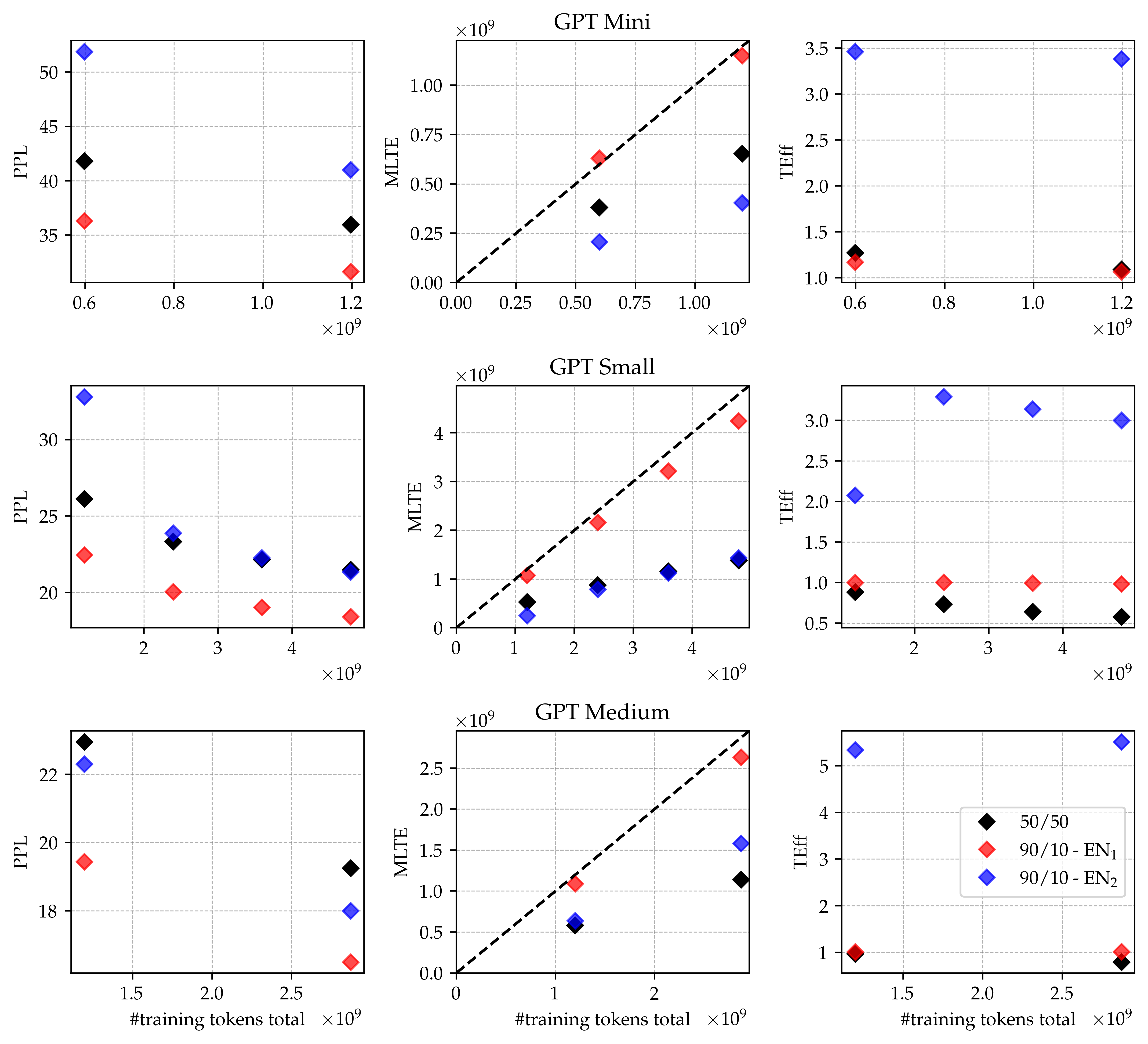}
    \caption{Performance with balanced and imbalanced $\english_1$ and $\english_2$ data for different configurations of model- and dataset size}
    \label{fig:fake_lang_more_data_larger_models}
\end{figure}

\begin{figure}[h]
    \centering
    \includegraphics[width=1.\linewidth]{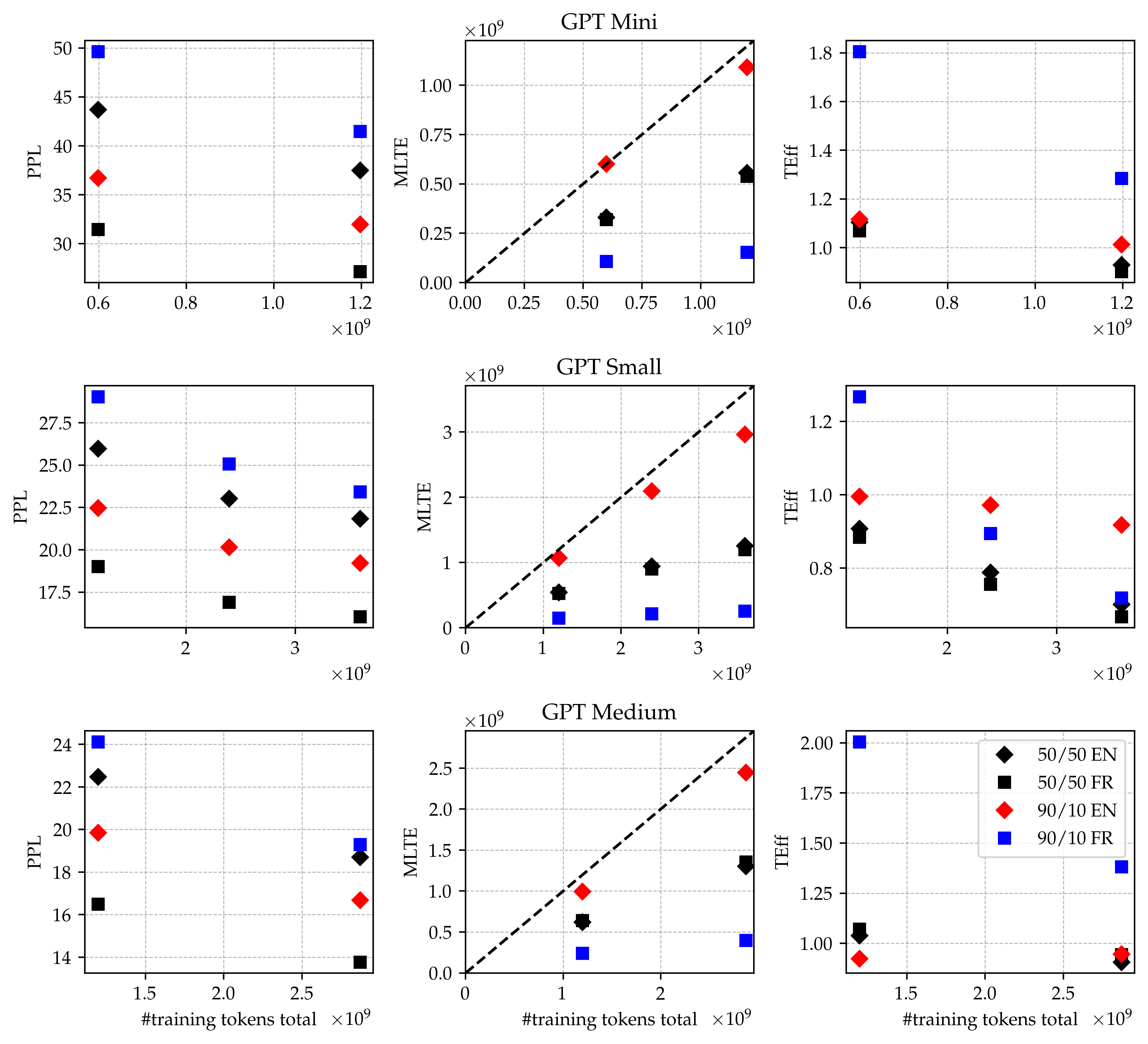}
    \caption{Performance with balanced and imbalanced $\english$ and $\french$ data for different configurations of model- and dataset size. Using anchored vocabulary.} 
    \label{fig:enfr_lang_more_data_larger_models}
\end{figure}

\cref{fig:fake_lang_more_data_larger_models} and \cref{fig:enfr_lang_more_data_larger_models} contain results for the full array of model- and dataset size combinations we ran for cloned languages and for English and French, respectively.

\clearpage
\section{Hidden State Similarity}
\label{app:hidden_sim}

Here, we compare the hidden states of our model when processing parallel sequences, both in cloned languages (see \cref{tab:en12_hidden_sim}) and in English and French (see \cref{tab:enfr_hidden_sim}). I.e., for a given trained model and parallel sequences $\words_a$ and $\words_b$, we first feed $\words_a$ through the model, then $\words_b$, and finally  compute the cosine similarities for the hidden states of pairs of corresponding tokens from $\words_a$ and $\words_b$ (see \cref{app:matching_hidden} for details on how these pairs are determined). We use 500 parallel sequences obtained from the Europarl parallel corpus \cite{koehn2005europarl}. For cloned languages, we observe that hidden states of the model trained under higher language imbalance generally have higher cosine similarity than the those of the model trained in a balanced setting. For English and French such a trend is less clear. Interestingly, however, an anchored vocabulary seems to lead to slightly higher similarities of the hidden states.

\begin{table*}[h]
\centering
\resizebox{\textwidth}{!}{%
\begin{tabular}{ccrrrrrrrrrrrr}
\toprule 
\multicolumn{2}{c}{Training Data} & \multicolumn{12}{c}{Layer} \\
\cmidrule(lr){1-2}\cmidrule(lr){3-14}
$p(\english_1)$ & $p(\english_2)$ & 1 & 2 & 3 & 4 & 5 &  6 & 7 & 8 & 9 & 10 & 11 & 12 \\
\midrule
50\% & 50\% & 0.55 & 0.79 & 0.83 & 0.88 & 0.85 & 0.83 & 0.78 & 0.66 & 0.56 & 0.46 & 0.25 & -0.21 \\
90\% & 10\% & 0.86 & 0.93 & 0.96 & 0.96 & 0.96 & 0.96 & 0.96 & 0.95 & 0.94 & 0.90 & 0.67 & 0.11 \\
\midrule
\multicolumn{2}{c}{$\Delta$} & 0.31 & 0.14 & 0.13 & 0.09 & 0.11 & 0.14 & 0.18 & 0.28 & 0.38 & 0.44 & 0.42 & 0.32 \\
\bottomrule
\end{tabular}%
}
\caption{Hidden states' cosine similarity when LM is fed equivalent inputs in cloned languages. Similarity is computed per token (i.e., comparing pairs of equivalent tokens).\looseness=-1}
\label{tab:en12_hidden_sim}
\end{table*}

\begin{table*}[h]
    \centering
    \resizebox{\textwidth}{!}{%
    \begin{tabular}{cccrrrrrrrrrrrr}
    \toprule 
    & \multicolumn{2}{c}{Training Data} & \multicolumn{12}{c}{Layer} \\
    \cmidrule(lr){2-3}\cmidrule(lr){4-15}
    & $p(\english)$ & $p(\french)$ & 1 & 2 & 3 & 4 & 5 &  6 & 7 & 8 & 9 & 10 & 11 & 12 \\
    \midrule
    \parbox[t]{2mm}{\multirow{3}{*}{\rotatebox[origin=c]{90}{\scalebox{0.85}{Disjoint}}}} & 
    50\% & 50\% & 0.68 & 0.80 & 0.84 & 0.88 & 0.86 & 0.84 & 0.80 & 0.75 & 0.62 & 0.53 & 0.34 & -0.15 \\
    & 90\% & 10\% & 0.71 & 0.83 & 0.88 & 0.87 & 0.86 & 0.84 & 0.81 & 0.74 & 0.69 & 0.57 & 0.40 & -0.17 \\
    \cmidrule(lr){2-15}
   & \multicolumn{2}{c}{$\Delta$} & 0.03 & 0.03 & 0.04 & -0.01 & 0.00 & 0.00 & 0.01 & 0.00 & 0.07 & 0.04 & 0.06 & -0.03 \\
    \midrule
    \parbox[t]{2mm}{\multirow{3}{*}{\rotatebox[origin=c]{90}{\scalebox{0.85}{Anchored}}}} & 
     50\% & 50\% & 0.73 & 0.84 & 0.88 & 0.91 & 0.89 & 0.88 & 0.85 & 0.78 & 0.71 & 0.61 & 0.36 & 0.10 \\
    & 90\% & 10\% & 0.78 & 0.87 & 0.89 & 0.91 & 0.89 & 0.87 & 0.84 & 0.77 & 0.72 & 0.63 & 0.28 & 0.06 \\
    \cmidrule(lr){2-15}
    & \multicolumn{2}{c}{$\Delta$} & 0.05 & 0.03 & 0.01 & 0.00 & 0.00 & -0.01 & 0.00 & -0.01 & 0.00 & 0.02 & -0.08 & -0.04 \\
    \bottomrule
    \end{tabular}%
    }
    \caption{Hidden states' cosine similarity for parallel inputs in $\english$ and $\french$ for anchored and disjoint vocabularies. We first match which tokens correspond to each other in the two languages, and then compare their representations (see \cref{app:matching_hidden}). \looseness=-1}
    \label{tab:enfr_hidden_sim}
    \vspace{-8pt}
\end{table*}

\clearpage
\section{Gradient Similarity}
\label{app:gradient_similarity}
\begin{figure}[h]
    \centering
    \includegraphics[width=1.\linewidth]{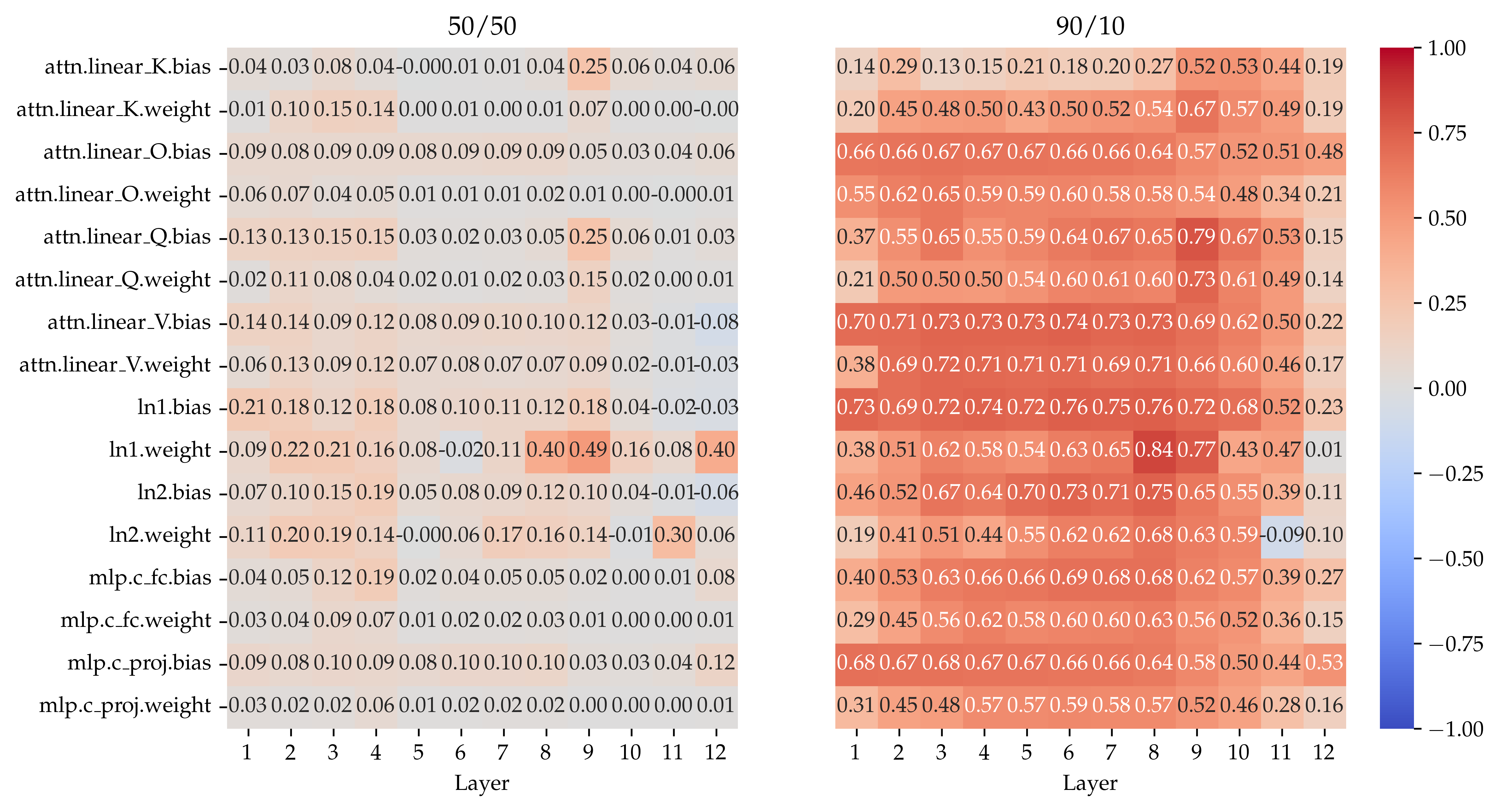}
    \caption{Similarity of gradients with respect to parallel sequences in $\english_1$ and $\english_2$ for models trained in balanced and imbalanced settings. Macro average for \sfrac{50}{50}: 0.07. Macro average for \sfrac{90}{10}: 0.53.}
    \label{fig:enfake_gradient_sims}
\end{figure}

\begin{figure}[h]
    \centering
    \includegraphics[width=1.\linewidth]{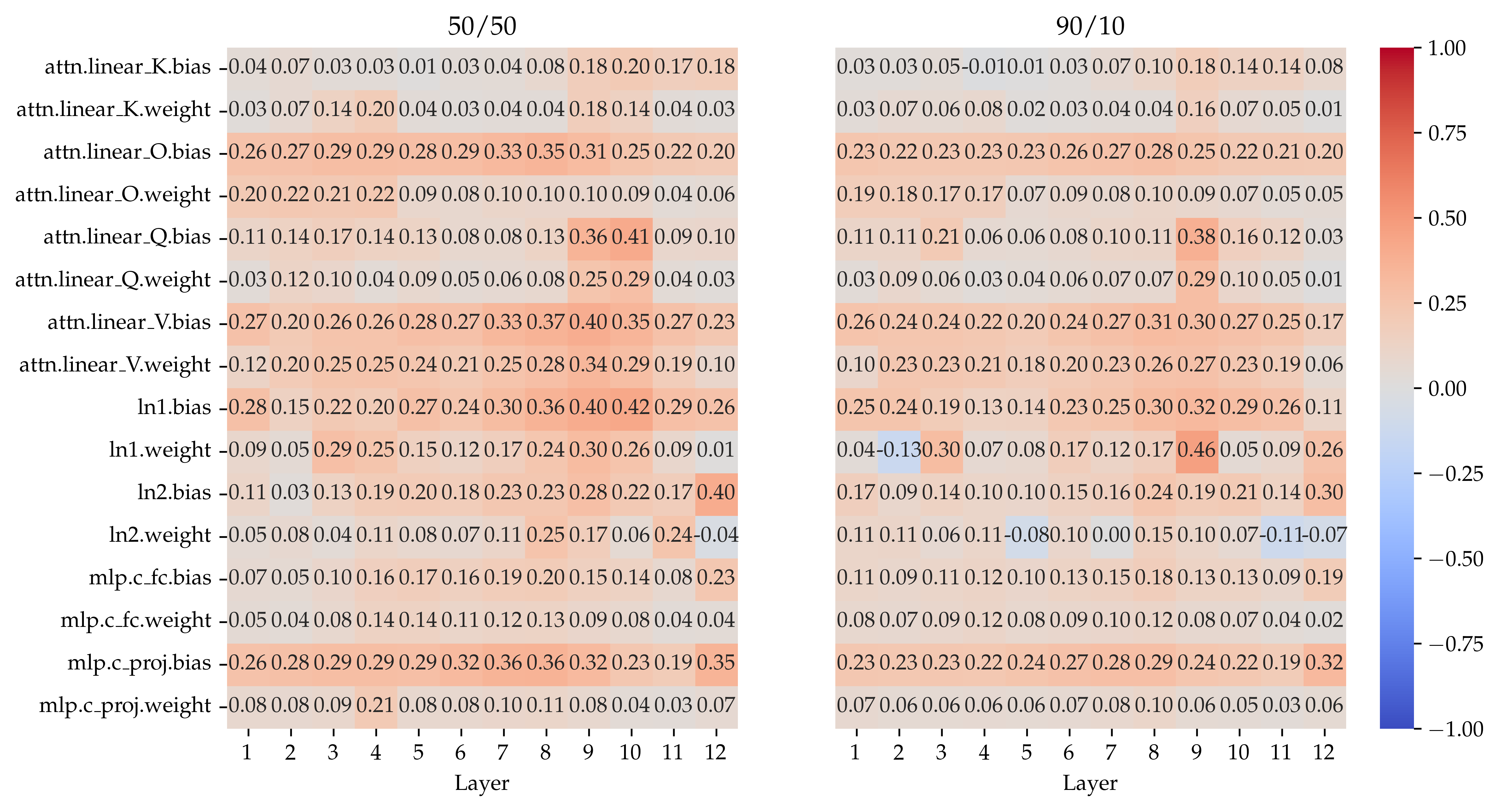}
    \caption{Similarity of gradients with respect to parallel sequences in $\english$ and $\french$ for models with anchored (i.e., merged) vocabulary, trained in balanced and imbalanced settings. Macro average for \sfrac{50}{50}: 0.17. Macro average for \sfrac{90}{10}: 0.14.}
    \label{fig:enfr_merged_gradient_sims}
\end{figure}

\begin{figure}[h]
    \centering
    \includegraphics[width=1.\linewidth]{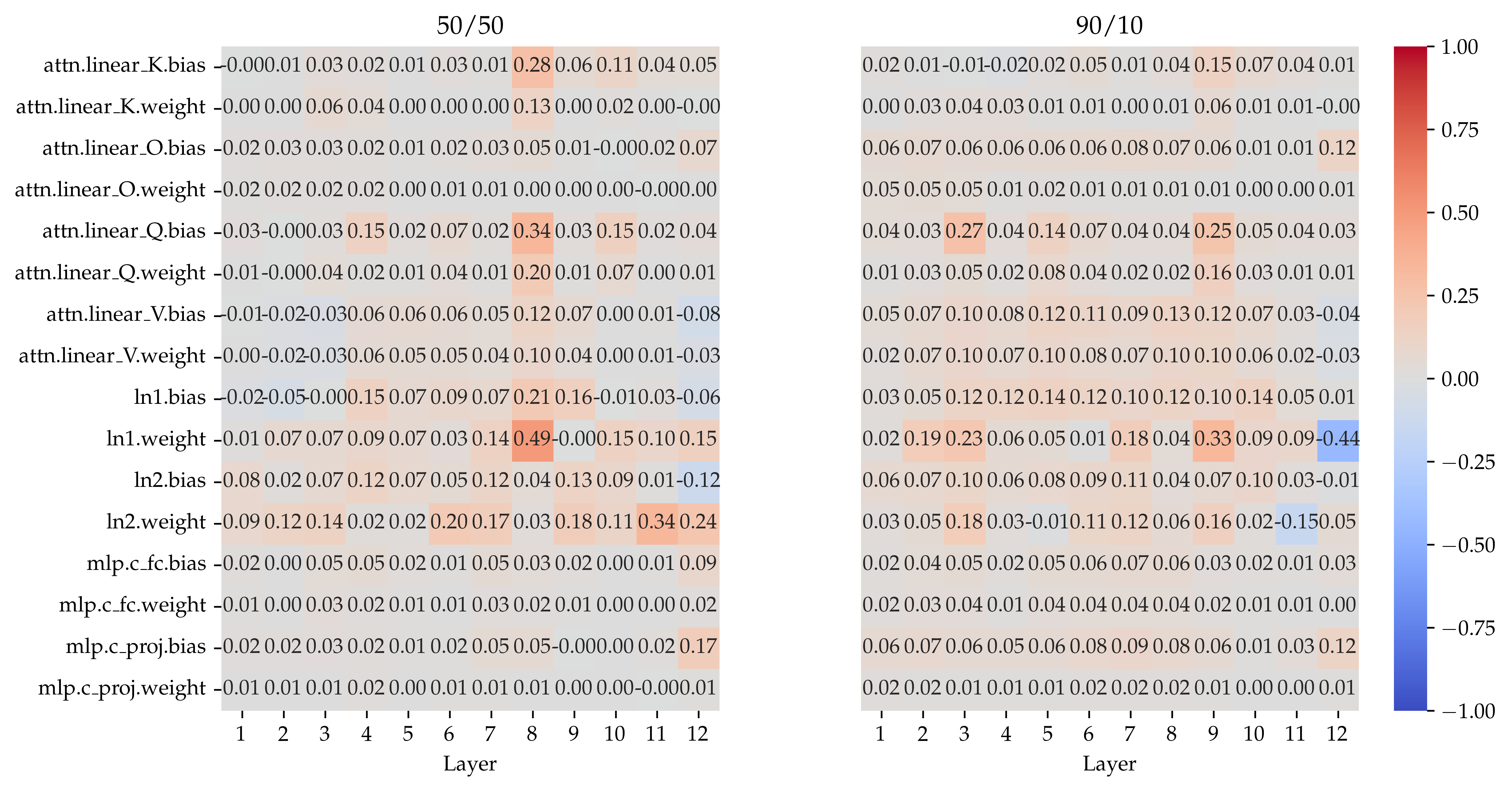}
    \caption{Similarity of gradients with respect to parallel sequences in $\english$ and $\french$ for models with disjoint vocabularies, trained in balanced and imbalanced settings. Macro average for \sfrac{50}{50}: 0.04. Macro average for \sfrac{90}{10}: 0.05.}
    \label{fig:enfr_unmerged_gradient_sims}
\end{figure}

Here, we compare the cosine similarity of trained models' gradients with respect to parallel sequences in two different (possibly cloned) languages. For cloned languages, the alignment between gradients is significantly higher for the model trained in the imbalanced \sfrac{90}{10} setting (see \cref{fig:enfake_gradient_sims}). 
For $\english$ and $\french$ data, this does not seem to be the case, whether the vocabulary is anchored (see \cref{fig:enfr_merged_gradient_sims}) or disjoint (see \cref{fig:enfr_unmerged_gradient_sims}). 
However, under the anchored vocabulary, the gradient similarities appear to be generally higher, suggesting better cross-lingual representation alignment.

\clearpage
\section{Matching Corresponding Tokens}
\label{app:matching_hidden}

\newcommand{\wordsenone}{\words_{\scaleto{\english_1}{5pt}}}
\newcommand{\wordsentwo}{\words_{\scaleto{\english_2}{5pt}}}
\newcommand{\tenone}{\word_{\scaleto{\english_1}{5pt}, t}}
\newcommand{\tentwo}{\word_{\scaleto{\english_2}{5pt}, t}}
\newcommand{\ten}{\word_{\scaleto{\english_{}}{5pt}, t}}
\newcommand{\tfr}{\word_{\scaleto{\french_{}}{5pt}, t'}}

In our experiments in \cref{sec:reallangs_imbalance}, we employ parallel sequences in different languages and compare both their hidden states' and their gradients' similarity.

When comparing gradients (see \cref{app:gradient_similarity}), we adopt a setup that is analogous to the training process as we aim to understand how one language might affect optimisation of the other: 
we compute gradients with respect to a full sequence in each language, and then compare these sequence-level aggregated gradients. 
Analogously, during training, gradient updates are also aggregated for entire sequences. 
(In fact, during training, these updates are also aggregated for an entire batch, but we use a batch size of 1 for this evaluation.)

However, when comparing hidden states, we compare the individual representations of corresponding tokens in the two sequences. 
We first compute the cosine similarity of each equivalent token pair, and only then average over the sequence dimension; this provides us with a more informative signal. 
For parallel sequences $\wordsenone \cloneequiv \wordsentwo$ in cloned languages, it is clear which token corresponds to which: At each given position $t$, we know that $\tenone \cloneequiv \tentwo$ so we can simply compare the hidden states position by position (see \cref{tab:en12_hidden_sim}).

Yet, this might not be the case for real languages $\english$ and $\french$, e.g., due to differing word order or tokenisation. To ensure that we still compare the hidden states of tokens that approximately correspond to each other in the respective languages, we match them based on their cosine similarity scores. Concretely, we create a bipartite graph where the nodes consist of the tokens of the two sequences. For every pair of tokens $\ten$ and $\tfr$ we add an edge which is weighed by the mean cosine similarity of their hidden states across all layers. We then compute a maximum weight full matching in this graph.\footnote{We compute the matching using the NetworkX \citep{hagberg2008exploring} implementation of the algorithm proposed by \citet{Karp:M78/67}.} Such a matching maximises the average similarity across all token pairs. Indeed, the resulting token pairs appear to approximately correspond to each other (see \cref{fig:word-alignment}). We can then compare the hidden states of these pairs (see \cref{tab:enfr_hidden_sim}).

Notably, the cosine similarities of hidden states of corresponding $\english$ and $\french$ tokens $\ten$ and $\tfr$ computed in this way generally appear to be slightly higher than for corresponding tokens $\tenone \cloneequiv \tentwo $ of cloned languages (compare \cref{tab:enfr_hidden_sim} (disjoint) and \cref{tab:en12_hidden_sim}). This might seem unexpected, given that $\tenone$ and $\tentwo$ are perfectly equivalent but $\ten$ and $\tfr$ are generally not. Could this be an artifact of the employed matching strategy which always maximises the average similarity, potentially matching tokens that have very high similarity but are completely unrelated? If this is the case, we should also obtain higher similarity scores in the cloned setting when using the described matching strategy instead of comparing position by position. 
After running this experiment, we find that using the matching strategy the similarities under the $\sfrac{50}{50}$ cloned language split are indeed marginally higher, although only in the last layers.
Under the $\sfrac{90}{10}$ split, however, we observe no notable changes. 
It thus seems that the proposed matching strategy does not artificially inflate similarity scores too strongly.

\usetikzlibrary{positioning}

\begin{figure}[h]
    \vspace{10pt}
    \centering

\def\SentenceOne{If, \_the, \_House, \_ag, rees, {,}, \_I, \_shall, \_do, \_as, \_Mr, \_Evans, \_has, \_suggested, {.}}
\def\SentenceTwo{Si, \_l, {'}, {Assemblée},  \_en, \_es', \_d, {'}, accord, {,}, \_je, \_ferai, \_comme, \_M, {.}, \_Ev, ans, \_l, {'}, a, \_sug, g, {éré}, {.}}

    \resizebox{\textwidth}{!}{%

\begin{tikzpicture}[wordbox/.style={draw, rectangle, inner sep=2pt, outer sep=0pt, anchor=west}]
      \node[wordbox] (word1-0) at (0,0) {\strut ...};
      \foreach [count=\xi from 1] \word in \SentenceOne {
        \node[wordbox, right=0pt of word1-\the\numexpr\xi-1\relax] (word1-\xi) {\strut\word};
      }

      \node[wordbox] (word2-0) at (0,-1.5) {\strut ...};
      \foreach [count=\xi from 1] \word in \SentenceTwo {
        \node[wordbox, right=0pt of word2-\the\numexpr\xi-1\relax] (word2-\xi) {\strut\word};
      }

      \node[wordbox, right=0pt of word1-15] (word1-inf) {\strut ...};
      \node[wordbox, right=0pt of word2-24] (word2-inf) {\strut ...};

      \draw[<->] (word1-1.south) -- (word2-1.north); %
      \draw[<->] (word1-2.south) -- (word2-2.north); %
      \draw[<->] (word1-0.south) -- (word2-3.north);  %
      \draw[<->] (word1-3.south) -- (word2-4.north); %
      \draw[<->] (word1-4.south) -- (word2-5.north); %
      \draw[<->] (word1-0.south) -- (word2-8.north); %
      \draw[<->] (word1-5.south) -- (word2-9.north); %
      \draw[<->] (word1-6.south) -- (word2-10.north); %
      \draw[<->] (word1-7.south) -- (word2-11.north); %
      \draw[<->] (word1-8.south) -- (word2-12.north); %
      \draw[<->] (word1-9.south) -- (word2-0.north); %
      \draw[<->] (word1-10.south) -- (word2-13.north); %
      \draw[<->] (word1-inf.south) -- (word2-14.north); %
      \draw[<->] (word1-11.south) -- (word2-15.north); %
      \draw[<->] (word1-12.south) -- (word2-19.north); %
      \draw[<->] (word1-13.south) -- (word2-20.north); %
      \draw[<->] (word1-14.south) -- (word2-23.north); %
      \draw[<->] (word1-15.south) -- (word2-24.north); %
    \end{tikzpicture}

}

    \caption{Computed matching for an example sentence using a model trained under \sfrac{50}{50} split with anchored vocabulary. Pointers to ``{...}'' denote a match with a token earlier or later in the sequence.}
    \label{fig:word-alignment}
\end{figure}
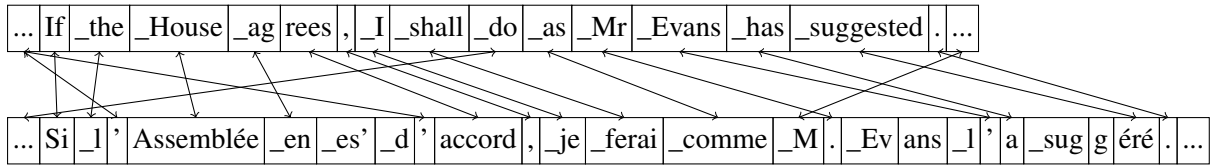

\end{document}